\documentclass[letterpaper, 10 pt, conference]{ieeeconf}  % Comment this line out
\IEEEoverridecommandlockouts                              % This command is only
\usepackage[nocompress]{cite}

\usepackage{graphics} % for pdf, bitmapped graphics files
\usepackage{epsfig} % for postscript graphics files
\usepackage{amsmath} % assumes amsmath package installed
\usepackage{amssymb}  % assumes amsmath package installed
\usepackage{mathtools}
\usepackage{bbold}
\usepackage{graphicx}
\usepackage{float}
\usepackage{subcaption}
\usepackage[bookmarks=false,hidelinks]{hyperref}
\usepackage{color}
\usepackage{verbatim}
\usepackage[algoruled,boxed,lined]{algorithm2e}
\newcommand{\signal}{X}
\DeclareMathOperator*{\argmax}{arg\,max}
\DeclareMathOperator*{\argmin}{arg\,min}

\title{\LARGE \bf
A Local Block Coordinate Descent Algorithm\\
for the Convolutional Sparse Coding Model
}

\author{Ev Zisselman$^*$, Jeremias Sulam$^\dagger$, Michael Elad$^\ddagger$%
         \thanks{$^*$ E. Zisselman is with the Department of Electrical-Engineering, Technion Israel Institute of Technology.}\\
         \thanks{$^\dagger$ J. Sulam is with the Department of Biomedical Engineering, Johns Hopkins University.}
         \thanks{$^\ddagger$ M. Elad is with the Department of Computer Science, Technion Israel Institute of Technology.}
}

\begin{document}

\maketitle
\thispagestyle{empty}

\pagestyle{empty}

%%%%%%%%%%%%%%%%%%%%%%%%%%%%%%%%%%%%%%%%%%%%%%%%%%%%%%%%%%%%%%%%%%%%%%%%%%%%%%%%
\begin{abstract}

The Convolutional Sparse Coding (CSC) model has recently gained considerable traction in the signal and image processing communities. By providing a global, yet tractable, model that operates on the whole image, the CSC was shown to overcome several limitations of the patch-based sparse model while achieving superior performance in various applications. Contemporary methods for pursuit and learning the CSC dictionary often rely on the Alternating Direction Method of Multipliers (ADMM) in the Fourier domain for the computational convenience of convolutions, while ignoring the local characterizations of the image. A recent work by Papyan et al. \cite{papyan2017convolutional} suggested the SBDL algorithm for the CSC, while operating locally on image patches. SBDL demonstrates better performance compared to the Fourier-based methods, albeit still relying on the ADMM.  

In this work we maintain the localized strategy of the SBDL, while proposing a new and much simpler approach based on the Block Coordinate Descent algorithm -- this method is termed Local Block Coordinate Descent (LoBCoD). Furthermore, we introduce a novel stochastic gradient descent version of LoBCoD for training the convolutional filters. The Stochastic-LoBCoD leverages the benefits of online learning, while being applicable to a single training image. We demonstrate the advantages of the proposed algorithms for image inpainting and multi-focus image fusion, achieving state-of-the-art results.

\end{abstract}

%%%%%%%%%%%%%%%%%%%%%%%%%%%%%%%%%%%%%%%%%%%%%%%%%%%%%%%%%%%%%%%%%%%%%%%%%%%%%%%%
\section{INTRODUCTION}

Sparse representation has been shown to be a very powerful model for many real-world signals, leading to impressive results in various restoration tasks such as denoising \cite{elad2006image}, deblurring \cite{dong2011image}, inpainting \cite{mairal2008sparse,elad2005simultaneous}, super-resolution \cite{yang2010image,dong2011image} and recognition \cite{wright2009robust}, to name a few. The core assumption of this model is that signals can be expressed as a linear combination of a few columns, also called atoms, taken from a matrix $\mathbf{D} \in \mathbb{R}^{N\times M}$ termed a dictionary. Concretely, for a signal $\signal \in \mathbb{R}^{N}$, the model assumption is that $\signal =\mathbf{D}\Gamma +V$, where $V$ is a noise vector with bounded energy $\|V\|_2<\epsilon$. This allows for a slight deviation from the model and/or may account for noise in the signal. The vector $\Gamma \in \mathbb{R}^{M}$ is the sparse representation of the signal, obtained by solving the following optimization problem \cite{EladBook,aharon2006rm}:

\begin{equation}\label{eq:1}
\hat{\Gamma}=\argmin_{\Gamma}\|\Gamma\|_0~~\text{s.t.}~~\|\signal-\mathbf{D}\Gamma\|_2<\epsilon,
\end{equation}
where $\|\cdot\|_0$ denotes the $l_0$ pseudo-norm that counts the number of non-zeros in the representation. Solving this optimization problem, known as the pursuit stage, is generally NP hard, but under certain conditions \cite{EladBook}, the solution of problem (\ref{eq:1}) can be approximated using greedy algorithms such as Orthogonal Matching Pursuit (OMP) \cite{chen1989orthogonal} or convex relaxation algorithms such as Basis Pursuit (BP) \cite{chen2001atomic}. Over the years, various methods have been proposed to adaptively learn the model parameters from real data. Such dictionary learning methods attempt to find $\mathbf{D}$ that best represents the set of signals at hand.
Prime examples are K-SVD \cite{aharon2006rm}, MOD \cite{engan1999method}, Double sparsity \cite{rubinstein2010double}, Online dictionary learning \cite{mairal2009online}, Trainlets \cite{sulam2016trainlets}, and more.

When dealing with high-dimensional signals, learning the dictionary suffers from the curse of dimensionality, and this process becomes computationally infeasible. To cope with this problem, many algorithms suggest training a local model on fully-overlapping patches taken from the signal $\signal$, and processing these patches independently. This patch-based dictionary learning technique has gained much popularity over the years due to its simplicity and high-performance \cite{elad2006image,dong2011image,mairal2008sparse,yang2010image}. Yet, patch-based approaches are known to be sub-optimal as they ignore the relations between neighboring patches \cite{sulam2015expected,romano2015patch}. 

An alternative approach to meet this challenge is posed by the Convolutional Sparse Coding (CSC) model. This model assumes that the signal can be represented as a superposition of a few local filters, convolved with sparse feature-maps. The CSC model handles the signal globally, and yet pursuit and dictionary learning are feasible due to the specific structure of the dictionary involved. This model has been shown to be useful in tackling some limitations of the patch-based model, and led to superior performance in several applications, such as super-resolution \cite{gu2015convolutional}, inpainting \cite{heide2015fast}, image separation \cite{papyan2017convolutional}, source separation \cite{liao2018monaural}, image fusion \cite{liu2016image} and audio processing \cite{grosse2012shift}, %albeit with some deficiencies.
albeit with room for improvement.

Contemporary CSC based algorithms often rely on the ADMM \cite{boyd2011distributed} formulation in order to extract the signal-representation of the model and train its corresponding filters. While the majority of works employ ADMM in the Fourier domain \cite{wohlberg2014efficient,heide2015fast,bristow2013fast}, a recent approach proposed by Papyan at el. \cite{papyan2017convolutional}, coined Slice Based Dictionary Learning (SBDL), adopts a local point of view and trains the filters in terms of only local computations in the signal domain. This local-global approach, and the decomposition that it induces, follows a recent work \cite{papyan2017working} that presented a novel theoretical analysis of the global CSC model, providing guarantees which stem from localized sparsity measures. The SBDL algorithm demonstrates state-of-the-art performance compared to the Fourier-based methods, while still relying on the ADMM algorithm. As such, this approach requires the introduction of $N$ auxiliary variables, increasing the memory requirements, it can only be deployed in a batch-learning mode, its convergence is questionable\footnote{While their pursuit method is provably converging, this is no longer the case when the dictionary is updated within the ADMM, as suggested in their work.}, and strongly depends on the ADMM parameter, which is application-dependent. 
% While the ADMM is guaranteed to converge to the global minimum of the pursuit formulation, such guarantees are lost when the dictionary is updated at every iteration of the SBDL.
% The reliance on the ADMM algorithm suffers from several limitations: firstly, the rate of convergence strongly depends on parameter-tuning in the pursuit stage, and secondly, there is no guarantee of convergence to a stationary point, a prevalent quality for stable algorithms.
%In this work, we maintain the localized strategy and propose an intuitive and easy-to-implement algorithm, based on the Block Coordinate Descent (BCD) algorithm, to solve the global problem in terms of local computations in the original domain, without the need for any parameter-tuning in the pursuit stage. We call this algorithm local block coordinate descent (LBCD). In addition, we introduce a new local stochastic gradient descent (LSGD) method for training the convolutional filters. This LSGD algorithm leverages the benefits of stochastic optimization on non-convex problems, while still being applicable to a single training-image. The LBCD algorithm combined with the LSGD algorithm shows faster convergence and achieves a better solution to the CSC problem compared to the previous ADMM-based methods.% also guaranteeing a descent direction update of the objective in each iteration and therefore converging globally to a stationary point. 

In this work we propose intuitive and easy-to-implement algorithms, based on the block coordinate descent approach, for solving the global pursuit and the CSC filter learning problems, all done with local computations in the original domain. The proposed algorithms operate without auxiliary variables nor extra parameters for tuning in the pursuit stage. We call this algorithm Local Block Coordinate Descent (LoBCoD). In addition, we introduce a stochastic gradient descent variant of LoBCoD for training the convolutional filters. This algorithm leverages the benefits of online learning, while being applicable even to a single training-image. The LoBCoD algorithm and its stochastic version show faster convergence and achieve a better solution to the CSC problem compared to the previous ADMM-based methods (global or local).

We should note that a very recent work by Moreau at el. \cite{moreau2018dicod} also proposes a coordinate descent based algorithm for the pursuit task in the CSC model. Their algorithm, like ours, operates locally and without necessitating additional parameters. However, the algorithm in \cite{moreau2018dicod} is restricted in two important ways compared to ours: (i) Their method is specifically tailored to 1D signals, and does not harness the 2D structure of the CSC model for images; (ii) their algorithm is limited to pursuit only, with no treatment for the filter learning.

The rest of this paper is organized as follows: Section \ref{sec:CSC} provides an overview of the CSC model and discusses previous methods. The proposed pursuit algorithm and its derivation are presented in Section \ref{sec:Proposed}. %In Section \ref{sec:LSGD}, we introduce the LSGD algorithm.
In Section \ref{sec:Dictionary Update} we discuss dictionary update methods and introduce the stochastic LoBCoD algorithm. We compare these methods with previously published approaches in section \ref{sec:Comparison}. Section \ref{sec:ImageProcessing} shows how our method can be employed to tackle the tasks of image inpainting and multi-focus image fusion, and later, in Section \ref{sec:Experiments}, we demonstrate this empirically. Section \ref{sec:Conclusions} concludes this work.
%In a very recent work by Moreau at el. \cite{moreau2018dicod} the authors established a coordinate descent based algorithm, which operated locally without necessitating additional parameters, and has the advantage of being computationally-distributive. However, their algorithm is limited to the pursuit of one dimensional signals and does not accommodate for training the filters.

\section{Convolutional sparse coding}\label{sec:CSC}
\begin{figure}[t] 
    \centering
     \includegraphics[scale=0.15]{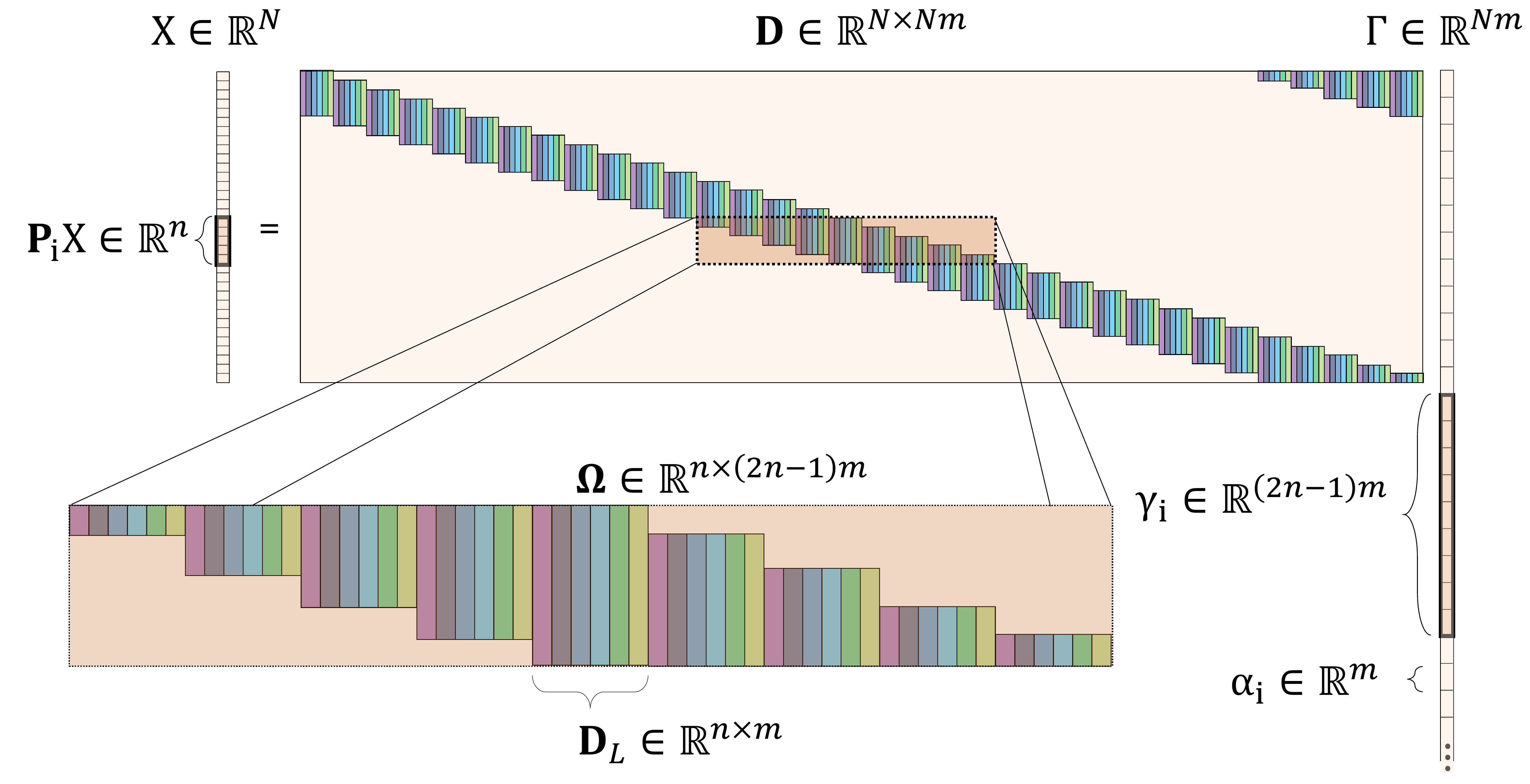}
     \caption{An illustration of the CSC model and its local components.}  
     \label{figure:a}
\end{figure}

The CSC model assumes that a signal\footnote{The description given here focuses on 1D signals for simplicity of the presentation. All our treatment applies to 2D (or higher dimensions) signals just as well.} $\signal \in\mathbb{R}^N$ can be represented by the sum of $m$ convolutions. These are built by feature maps $\{Z_i\}_{i=1}^m$, each of length of the original signal $N$, convolved with $m$ small support filters $\{d_i\}_{i=1}^m$ of length $n\ll N$. In the dictionary learning problem, one minimizes the following cost function over both the filters and the feature maps\footnote{Throughout the subsequent derivations we assume that the filters are normalized to a unit $l_2$-norm.}:
%The feature maps and the filters are learned together by solving the following optimization problem \footnote{Throughout the subsequent derivations we assume that the filters are normalized to a unit $l_2$ norm.}
\begin{equation}\label{eq:2}
\begin{aligned}
\min_{d_i,Z_i}\frac{1}{2}\|\signal-\sum_{i=1}^{m}d_i\ast Z_i\|_2^2+\lambda\sum_{i=1}^{m}\|Z_i \|_1.\\
%\text{s.t}~~\|d_i \|_2 =1.
\end{aligned}
\end{equation}
Given the filters, the above problem becomes the CSC pursuit task of finding the representations $\{Z_i\}_{i=1}^m$. Consider a global dictionary $\mathbf{D}$ to be the concatenation of $m$ banded circulant matrices, where each matrix represents a convolution with one filter $d_i$. By permuting its columns, the global dictionary $\mathbf{D}$ consists of all shifted versions of a local dictionary $\mathbf{D}_L$ of size $n\times m$, containing the filters $\{d_i\}_{i=1}^m$ as its columns, and the global sparse vector $\Gamma$ is simply the interlaced concatenation of all the feature maps $\{Z_i\}_{i=1}^m$. Such a structure is depicted in Fig. \ref{figure:a}. Using the above formulation, the convolutional dictionary learning problem (\ref{eq:2}) can be rewritten as
\begin{equation}\label{eq:3}
\min_{\mathbf{D},\Gamma}\frac{1}{2}\|\signal-\mathbf{D}\Gamma\|_2^2+\lambda\|\Gamma\|_1.
\end{equation}
\noindent  Similar to our earlier comment, when $\mathbf{D}$ is known, we obtain the CSC pursuit problem, defined as 
\begin{equation}\label{eq:3a}
\min_{\Gamma}\frac{1}{2}\|\signal-\mathbf{D}\Gamma\|_2^2+\lambda\|\Gamma\|_1.
\end{equation}
%Assuming a global signal $x$ to be represented under the CSC model as $D\Gamma$, Fig. \ref{figure:a} describes such a global signal, its corresponding dictionary $D\in\mathbb{R}^{N\times mN}$, and its sparse representation $\Gamma\in\mathbb{R}^{mN}$. 
Herein, we review some of the  definitions from \cite{papyan2017working} as they will serve us later for the description of our algorithms.

The global sparse vector $\Gamma$ can be broken into $N$ non-overlapping $m$ dimensional local vectors $\alpha_i$, referred to as \textit{needles}. This way, one can express the global vector $\signal$ as $\signal=\sum_{i=1}^N\mathbf{P}_i^T\mathbf{D}_L\alpha_i$, where $\mathbf{P}_i^T\in\mathbb{R}^{N\times n}$ is the operator that positions $\mathbf{D}_L\alpha_i$ in the i-th location and pads the rest of the entries with zeros. 
On the other hand, a patch $\mathbf{P}_i\signal=\mathbf{P}_i\mathbf{D}\Gamma$ taken from the signal $\signal$ equals to $\boldsymbol\Omega\gamma_i$ (see Fig. \ref{figure:a}), where $\boldsymbol\Omega\in\mathbb{R}^{n\times (2n-1)m}$ is a \textit{stripe} dictionary containing $\mathbf{D}_L$ in its center, and $\gamma_i$ is the \textit{stripe} vector containing the local vector $\alpha_i$ in its center. In other words, a stripe $\gamma_i$ is the sparse vector that codes all the content in the patch $\mathbf{P}_i\signal$, whereas a needle $\alpha_i$ only codes part of the information within it.

The theoretical work in \cite{papyan2017working} suggested an analysis of the CSC global model, augmented by a localized sparsity measure. Specifically, this work showed that if all the stripes $\gamma_i$ are sparse, the solution to the convolutional sparse pursuit problem is unique and can be recovered by a greedy algorithm, such as the OMP \cite{chen1989orthogonal}, or a convex relaxation algorithm such as the BP \cite{chen2001atomic}. They extended the analysis to a noisy regime, showing that under similar sparsity assumptions, the pursuit algorithms for the global formulation are also stable. Inspired by this analysis, herein we maintain such a local-global decomposition and propose a global algorithm that operates locally on image patches. Prior to describing our algorithm, we turn to review a closely related work, the SBDL algorithm \cite{papyan2017convolutional}.
%Herein, we adhere to the local-global relation and propose a global algorithm that operates locally on image patches. Prior to describing our algorithm, we turn to review a previous related work, the SBDL algorithm \cite{papyan2017convolutional}.     

Equipped with the above definitions and the separability of the \textit{$l_1$} norm, we can express the global CSC problem in terms of the local sparse vectors $\alpha_i$ and the local dictionary $\mathbf{D}_L$ by
\begin{equation}\label{eq:4}
\min_{\mathbf{D}_L,\{\alpha_i\}}\frac{1}{2}\|\signal-\sum_{i=1}^N\mathbf{P}_i^T\mathbf{D}_L\alpha_i\|_2^2+\lambda\sum_{i=1}^N\|\alpha_i\|_1.
\end{equation}
To solve this problem, the Slice Based Dictionary Learning (SBDL) algorithm \cite{papyan2017convolutional} adopts a variable splitting approach. By denoting $\mathbf{D}_L\alpha_i$ as the i-th slice $s_i$ and writing the global signal in terms of the slices $\signal=\sum_{i=1}^N\mathbf{P}_i^Ts_i$, the above can be written as the following constrained minimization problem,
\begin{equation}\label{eq:5}
\begin{aligned}
\min_{\mathbf{D}_L,\{\alpha_i\},\{s_i\}}\frac{1}{2}\|\signal-\sum_{i=1}^N\mathbf{P}_i^Ts_i\|_2^2+\lambda\sum_{i=1}^N\|\alpha_i\|_1\\
~~ \text{s.t}~~ \{s_i=\mathbf{D}_L\alpha\}_{i=1}^N.~~ ~~ ~~ ~~ ~~ ~~ 
\end{aligned}
\end{equation}
For solving (\ref{eq:5}), the SBDL employs the ADMM algorithm \cite{boyd2011distributed}, which translates the constraints to penalties, and minimizes the following augmented Lagrangian problem
\begin{equation}\label{eq:6}
\begin{aligned}
\min_{\stackrel{\mathbf{D}_L,\{\alpha_i\},}{\{s_i\},\{u_i\}}}
\frac{1}{2}\|\signal-\sum_{i=1}^N\mathbf{P}_i^Ts_i\|_2^2~~ ~~ ~~ ~~ ~~ ~~ ~~ ~~ ~~ ~~ ~~  \\
+\sum_{i=1}^N(\lambda\|\alpha_i\|_1+\frac{\rho}{2}\|s_i-\mathbf{D}_L\alpha_i+u_i\|_2^2),
\end{aligned}
\end{equation}
by optimizing with respect to every set of variables $\{\alpha_i\}_{i=1}^N,\{s_i\}_{i=1}^N,\{u_i\}_{i=1}^N$ sequentially. Here $\{u_i\}_{i=1}^N$ denote the dual variables of the ADMM formulation.
The minimization with respect to the needles $\{\alpha_i\}_{i=1}^N$ is separable and boils down to a traditional patch-based BP formulation. In particular, the work in \cite{papyan2017convolutional} employed the batch-LARS \cite{efron2004least} algorithm for this stage,
%was resolved using the LARS algorithm, 
whereas the minimization with respect to the slices $\{s_i\}_{i=1}^N$ amounts to a simple Least Squares problem. The SBDL also accommodates learning of the filters, i.e the local dictionary $\mathbf{D}_L$, using any patch-based dictionary learning algorithm such as the K-SVD \cite{aharon2006rm} or the MOD \cite{engan1999method}. Note, however, that in adding this part within the ADMM, all the convergence  guarantees are lost.      

In this work, unlike the above variable splitting approach, we leverage the block coordinate descent algorithm to update the local sparse vectors (needles) without defining any additional variables. In this manner, we avoid the need of tuning extra parameters, conserve memory, all while demonstrating superior performance. We also propose a dictionary update scheme, which can operate both in batch and online modes.

\section{Proposed Method: CSC Pursuit}\label{sec:Proposed}
\subsection{Local Block Coordinate Descent}\label{subsec:Pursuit}
In this section we focus on the pursuit of the representations, leaving the study of updating the dictionary for Section \ref{sec:Dictionary Update}.
The convolutional sparse coding problem presented in the previous section is solved by minimizing the global objective of Equation (\ref{eq:3a}). In this paper, we adopt a local strategy and split the global sparse vector $\Gamma$ into local vectors, needles, as described in Equation (\ref{eq:4}). However, rather than optimizing with respect to all the needles together, we can treat each needle $\alpha_i$ as a block of coordinates, taken from the global vector $\Gamma$, and optimize with respect to each such block separately, and sequentially. Consequently, the update rule of each needle can be written as
\begin{equation}\label{eq:7}
\begin{aligned}
\min_{\alpha_i}\frac{1}{2}\|(\signal-\sum^N_{\stackrel{j=1}{j\neq i}} \mathbf{P}_j^T\mathbf{D}_L\alpha_j)-\mathbf{P}_i^T\mathbf{D}_L\alpha_i\|_2^2+\lambda\|\alpha_i\|_1.
\end{aligned}
\end{equation}
By defining $R_i=(\signal-\sum_{\stackrel{j=1}{j\neq i}}^N\mathbf{P}_j^T\mathbf{D}_L\alpha_j)$ as the residual image without the contribution of the needle $\alpha_i$, we can rewrite Equation (\ref{eq:7}) as
\begin{equation}\label{eq:8}
\min_{\alpha_i}\frac{1}{2}\|R_i-\mathbf{P}_i^T\mathbf{D}_L\alpha_i\|_2^2+\lambda\|\alpha_i\|_1.
\end{equation}
While the above minimization involves global variables, such as the residual $R_i$, one can show (see Appendix A) that this can be decomposed into an equivalent and local problem:
%While the above update rule optimizes with respect to the local sparse vector $\alpha_i$, it still involves global variables (such as the residual). On the other hand, one can note that the update rule of the needle $\alpha_i$ is affected only by the corresponding i-th patch in the original signal $x$, and by the immediate needles of $\alpha_i$ (the needles of the stripe vector $\gamma_i$, where $\alpha_i$ is in its center). In other words, by denoting $p_i$ as the patch which fully contains the slice $s_i=D_L\alpha_i$, the above update rule is affected only by the corresponding patch $p_i$ of the original signal $x$, and of the reconstructed signal $\widehat{x}=\Gamma x$. Therefore, instead of solving the high dimensional problem \ref{eq:8}, we can solve the corresponding reduced optimization problem, iteratively:
\begin{equation}\label{eq:9}
\min_{\alpha_i}\frac{1}{2}\|\mathbf{P}_i R_i-\mathbf{D}_L\alpha_i\|_2^2+\lambda\|\alpha_i\|_1.
\end{equation}
This follows from the observation that the update rule of the needle $\alpha_i$ is effected only by pixels belonging to the corresponding patch $\mathbf{P}_i R_i$ (the part that fully overlaps with the slice $s_i=\mathbf{D}_L\alpha_i$). For more details, we refer the reader to Appendix A.

\begin{algorithm}
\caption{The LoBCoD basic pursuit algorithm}\label{algorithm_1}

    \textbf{Input:} signal $\signal$, dictionary $\mathbf{D}_L$, initial needles $\{\alpha_i^0\}_{i=1}^N$\\
	\textbf{Output:} needles $\{\alpha_i\}_{i=1}^N$\\
    \textbf{Initialization:} ~~$R = \signal$, $\widehat{\signal} =0$, $k = 0$\\
    \vspace{0.25cm}
   	\While {\textit{not converged}}{
     $k=k+1$ \\
    	\For{\textit{i = 1:N}}{
        	 Computation of the local residual:
             \begin{equation}\nonumber R_i = R + \mathbf{P}_i^T\mathbf{D}_L\alpha_i^{k-1}\end{equation}
			 
             Local sparse pursuit:
            \begin{equation}\nonumber \alpha_i^k=\argmin_{\alpha_i}\frac{1}{2}\|\mathbf{P}_i R_i - \mathbf{D}_L\alpha_i\|_2^2+\lambda\|\alpha_i\|_1\end{equation}
            
             Update of the reconstructed signal:
            \begin{equation}\nonumber \widehat{\signal} = \widehat{\signal} + \mathbf{P}_i^T\mathbf{D}_L(\alpha_i^k-\alpha_i^{k-1})\end{equation}
            
             Update of the residual signal:
            \begin{equation}\nonumber R = \signal-\widehat{\signal}\end{equation}
            
       }
    }
\end{algorithm}

The main idea of the block coordinate descent algorithm is that every step minimizes the overall penalty w.r.t. a certain block of coordinates, while the other ones are set to their most updated values. Following this idea, every local pursuit stage (\ref{eq:9}) proceeds by updating the global reconstructed signal $\widehat{\signal}$ and the global residual $R = \signal - \widehat{\signal}$, as a preprocessing stage to the consecutive stage that updates the next needle, based on the most updated values of the previous needles. This pursuit algorithm is summarized in Algorithm \ref{algorithm_1}.  

 \begin{algorithm}
 \caption{The stochastic LoBCoD pursuit and dictionary learning algorithm}\label{algorithm_2}

    \textbf{Input:} signal $\signal$, initial dictionary $\mathbf{D}_L$, initial needles $\{\alpha_i^0\}_{i=1}^N$\\
 	\textbf{Output:} needles $\{\alpha_i\}_{i=1}^N$, the trained dictionary $\mathbf{D}_L$ \\
    \textbf{Initialization:}~~~$R = \signal$, $\widehat{\signal} =0$, $k = 0$\\
    \vspace{0.25cm}
    \While {\textit{not converged}}{
     $k=k+1$ \\
     \For{\textit{ j = 1:$n$}}{
        Computation of the residual:
 	   \begin{equation}\nonumber R_j = R + \sum_{i \in L_j}\mathbf{P}_i^T  \mathbf{D}_L\alpha_i^{k-1} \end{equation}
			 
        Sparse pursuit: $\forall i\in L_j$ (in parallel)
        \begin{equation}\nonumber \alpha_i^k=\argmin_{\alpha_i}\frac{1}{2}\|\mathbf{P}_i R_j - \mathbf{D}_L\alpha_i\|_2^2+\lambda\|\alpha_i\|_1\end{equation}            
        
        Computation of the reconstructed signal:
        \begin{equation}\nonumber \widehat{\signal} = \widehat{\signal} + \sum_{i \in L_j}\mathbf{P}_i^T \mathbf{D}_L(\alpha_i^k-\alpha_i^{k-1})\end{equation}       
        
       Computation of the residual signal:
        \begin{equation}\nonumber R = \signal-\widehat{\signal}\end{equation}          

    \fbox{
    \parbox{0.377\textwidth}{
       Computation of the gradient w.r.t $\mathbf{D}_L$:
        \begin{equation}\nonumber \nabla_{\mathbf{D}_L} =-\sum_{i \in L_j}\mathbf{P}_i R\cdotp(\alpha_i^k)^T\end{equation}           
       
       Dictionary update:
        \begin{equation}\nonumber \mathbf{D}_L =\mathcal{P}_{\mathbb{1}}[\mathbf{D}_L-\eta\nabla_{\mathbf{D}_L}]\end{equation} 
      }
    }
     }
   }
\end{algorithm}
An important insight is that needles that have no footprint overlap in the image can be updated efficiently in parallel in the above algorithm without changing the algorithm's outcome. This enables employing efficient batch-implementations of the LARS algorithm. Alternatively, the calculation can be distributed across multiple processors to gain a significant speedup in performance. To formalize these observations, we define the \textit{layer} $L_i$ as the set of needles that have no induced overlap in the image. We sweep through these layers and update their respective needles in parallel, followed by updating the global reconstructed signal $\widehat{\signal}$ and the global residual $R$. This way, the number of the layers imposes the number of the inner iterations, which will determine the complexity of our final algorithm. As opposed to Algorithm \ref{algorithm_1} that has $N$ inner iterations since it iterates trough the needles, here the number of the inner iterations depends only on the patch size; for $\sqrt{n}\times \sqrt{n}$ patches, the number of layers is $n$. This parallelized pursuit algorithm is presented in Algorithm \ref{algorithm_2}.

Note that this algorithm can clearly be extended to iterate over multiple signals, but for the sake of brevity we assume that the data corresponds to an individual signal $\signal$.

\subsection{Boundary Conditions and Initialization}\label{subsec:Boundaries}
In the formulation of the CSC model, as shown in Fig. \ref{figure:a}, we assumed that the dictionary is comprised of a set of banded circulant matrices, which impose a circulant boundary conditions on the signals. In practice, however, signals and images do not exhibit circulant boundary behavior. Therefore, our model incorporates a preemptive treatment of the boundaries. We adopt a similar approach to \cite{papyan2017convolutional}, in which the signal boundaries are padded with $n-1$ elements prior to decomposing it with the model. At the end of the process, we discard the added padding by cropping the $n-1$ boundary elements from the reconstructed signal and from the resulting feature maps (sparse representation). 

%In addition, one should note that although we have presented the above derivations for 1D vectors for simplicity, all of our discussion applies directly to two-dimensional images. This can be performed by replacing the Circulant matrices with Block-Circulant Circulant-Block (BCCB) matrices.

Another beneficial preprocess step is needles initialization. A good initialization would equally spread the contribution of the needles towards signal reconstruction. With that goal, we set the initial value of each needle $\alpha_i$ to be the sparse representation of $\frac{1}{n}\mathbf{P}_i\signal$, i.e its relative portion of the corresponding patch. This can be done by solving the following local pursuit for every needle:
\begin{equation}\label{eq:99}
\alpha_i^0=\argmin_{\alpha_i}\frac{1}{2}\|\frac{1}{n}\mathbf{P}_i\signal-\mathbf{D}_L\alpha_i\|_2^2+\lambda\|\alpha_i\|_1,
\end{equation}
as a preprocess stage of our algorithm. 

\section{CSC Dictionary Learning}\label{sec:Dictionary Update}
When addressing the question of learning the CSC filters, the common strategy is to alternate between sparse-coding and dictionary update steps for a fixed number of iterations. The dictionary update step aims to find the minimum of the quadratic term of Equation (\ref{eq:4}) subject to the constraint of normalized dictionary columns:
\begin{equation}\label{eq:10}
\begin{aligned}
\min_{\mathbf{D}_L}\frac{1}{2}\|\signal-\sum_{i=1}^N\mathbf{P}_i^T\mathbf{D}_L\alpha_i\|_2^2 ~~ ~~\\
~~ ~~ ~~ ~~ ~~ ~~ ~~ \text{s.t}~\{\|d_i\|_2 = 1\}_{i=1}^m.~~ ~~ ~~
\end{aligned}
\end{equation}
One can do so in a batch manner which requires access to the entire data set at every iteration, or in an online (stochastic) manner that enables access to only small part of the dataset at every update step. This way it is also applicable for streaming data scenarios, when the probability distribution of the data changes over time. 

\subsection{Batch Update}\label{sec:Batch}
Usually, for offline applications, when the whole data set is given, the batch approach is generally simpler, and thus we start with its description. The typical approach is to alternate between sparse coding (\ref{eq:3a}) and dictionary update (\ref{eq:10}) phases. For the latter, solving problem (\ref{eq:10}) requires finding the optimum $\mathbf{D}_L$ that satisfies the normalization constraint. One can find this optimal solution using projected steepest descent: perform steepest descent with a small step size and project the solution to the constraint set after each iteration, until convergence.   
To that end, the gradient of the quadratic term in Equation (\ref{eq:10}) w.r.t. $\mathbf{D}_L$ is\footnote{The full derivation for the gradient can be found in Appendix B.}:
%Computing the gradient of the quadratic term of Equation (\ref{eq:10}) w.r.t $\mathbf{D}_L$, we get:
\begin{equation}\label{eq:11} 
\nabla_{\mathbf{D}_L} =-\sum_{i=1}^N\mathbf{P}_i(\signal-\widehat{\signal})\cdotp\alpha_i^T.
\end{equation}
The final update step for the local dictionary $\mathbf{D}_L$ is obtained by advancing in the direction of this gradient (\ref{eq:11}) and normalizing the columns of the resulting $\mathbf{D}_L$ in each iteration, until convergence.  

This batch dictionary update rule follows the line of thought of the MOD algorithm \cite{engan1999method}, and thus improves the solution in each step.
However, it exhibits a very slow convergence rate since each dictionary update can be performed only after finishing the entire sparse coding (pursuit) stage, which is markedly inefficient, as the pursuit is the most time consuming part of the algorithm.
%Although this batch dictionary update rule is very simple and intuitive, it leads to a very slow convergence rate since each dictionary update step can be performed only after finishing the entire sparse coding (pursuit) stage, which is significantly inefficient as the sparse coding is the most time consuming part of the algorithm. 
This brings us to the Stochastic-LoBCoD alternative. 
%intolerably large matrix-multiplications and matrix-inversion. More severely, each dictionary update step can be performed only after finishing the whole sparse coding stage, which is inefficient. Therefore, the practicality of this algorithm is limited due to a very slow convergence rate. In the next subsection we present the stochastic LoBCoD algorithm which addresses these issues.      

\subsection{Local Stochastic Gradient Descent Approach}\label{sec:LSGD}
The traditional Stochastic Gradient Descent (SGD) approach restricts the computation of the gradient to a subset of the data and advances in the direction of this noisy gradient with every update step. Building upon this concept and the fact that Equation (\ref{eq:11}) reveals a separable gradient w.r.t the patches and their corresponding needles, we can update the dictionary in a stochastic manner. Rather than concluding the entire pursuit stage and then advancing in the direction of the global gradient, we can take a small step size $\eta$ and update the dictionary after finding the sparse representation of only a small group of needles. According to Section \ref{sec:Proposed}, every iteration updates a group of needles, referred to as a layer $L_i$, which in turn could now serve to update the dictionary. This way, our algorithm convergences faster and adopts the stochastic behavior of the SGD while still operating on a single image. 

The filters should be normalized after every dictionary update by projecting them onto the $l_2$ unit ball. Here, due to the choice of small step size, we simply normalize the atoms after every dictionary update:
\begin{equation} 
\nonumber \mathbf{D}_L =\mathcal{P}_{\mathbb{1}}[\mathbf{D}_L-\eta\nabla_{\mathbf{D}_L}].
\end{equation}
Where $\mathcal{P}_{\mathbb{1}}[\cdot]$ denotes the operator that projects the dictionary atoms onto the unit ball. The final algorithm that incorporates the dictionary update is summarized in Algorithm \ref{algorithm_2}.

Note that, although this dictionary update rule introduces an extra parameter (the step size $\eta$), determining its value is rather intuitive and can be performed automatically by setting it to $1-2\%$ of the norm of the gradient. Furthermore, this update rule may also leverage any stochastic optimization algorithm such as Momentum, Adagrad, Adadelta, Adam \cite{ruder2016overview} etc., with their authors' recommended parameter values. This choice of parameter setting is sufficient, as will be demonstrated empirically in Section \ref{sec:Experiments}. In the rest of this work we will use this dictionary update rule, as it shows superior results. 

\section{Relation to Other Methods}\label{sec:Comparison}
In this section we evaluate the proposed approach and describe its advantages over the Fourier and ADMM based methods.

\begin{enumerate}
\item \textbf{Parallel computation:} Our algorithm is trivial to parallelize efficiently across multiple processors by virtue of operating directly on the image patches. One can split the computation between $N/n$ processors, in correspondence with the number of the needles in every layer, and perform the local sparse pursuit stage in-parallel for all the needles in the same layer. At the end of this stage, and in preparation for computing the next layer, every processor needs to pass its new local sparse result solely to its neighboring processors, i.e. those which act on common patches. This way, each processor waits only for its neighbors to finish their calculations, and is unhindered by processors that target farther regions of the image. This path of computing maintains its efficacy even in case where the processors differ in their capacity, or in case of large variation in the local complexity of the image. The global aggregation is performed only once, if needed, at the end of the algorithm to produce the reconstructed image. Such a parallel approach cannot be directly applied with ADMM based algorithms that use auxiliary variables, or with Fourier-based methods that lose the relation to the local patches of the image.

\item \textbf{Online learning:} The proposed algorithm, due to its local stochastic manner, can work in a streaming mode, where the probability distribution of the patches varies over time. It is unclear how to adopt such an approach in the Fourier-based methods \cite{heide2015fast,wohlberg2014efficient}, considering the global nature of the Fourier domain, or in the SBDL algorithm, \cite{papyan2017convolutional} considering its use of auxiliary variables.
Another aspect of this advantage is our ability to run in an online manner, even for a single input image. This stands in sharp contrast to other recent online methods \cite{wang2018scalable,liu2017online} which allow for online training but only in the case of streaming images. Other approaches took a step further and proposed partitioning the image into smaller sub-images \cite{liu2018first}, but this is still far from our approach, which can stochastically estimate the gradient for each needle.
\item \textbf{Parameter free:} Contrary to ADMM-based approaches, our algorithm is unhindered by cumbersome manual parameter-tuning at the pursuit stage. Moreover, it benefits from an intuitively tuned parameter (the step size $\eta$) in the dictionary learning stage, as described as Section \ref{sec:Dictionary Update}.

\item \textbf{Memory efficient:} Our algorithm has better storage complexity compare to the ADMM-based approaches \cite{papyan2017convolutional,heide2015fast} since the update of the sparse vector is performed in-place and does not require any auxiliary variables. For example, the SBDL \cite{papyan2017convolutional} requires $O(N)$ auxiliary variables for every patch in the image, each of these variables is of patch-size $\sqrt{n}\times \sqrt{n}$, thus this methods requires $O(Nn)$ extra memory compare to our algorithm.

\item \textbf{Adaptive local complexity:} As opposed to the Fourier-oriented algorithms, our algorithm is attuned to the local properties of the signal. As such, it can be easily modified to allow an adaptive number of non-zeros in different regions of the global signal.

\end{enumerate}

\begin{table}[t]
	\centering
	\begin{tabular}{|p{0.04\textwidth}|c|}
		\hline
		{\scriptsize Method}  & {\scriptsize Time Complexity}                                                 \\ \hline
		{\scriptsize  \begin{tabular}[c]{@{}c@{}} \cite{liu2018first} \\ (Sparse)\end{tabular}} & {\scriptsize $\underbrace{ \color{red}\boldsymbol{q} \boldsymbol{I} \boldsymbol{m} \boldsymbol{N} \boldsymbol{log(N)}}_{\text{T-pursuit}} ~~~ + \underbrace{ I N m k}_{\text{SGD using sparse matrix}}$ }\\ \hline
		{\scriptsize  \begin{tabular}[c]{@{}c@{}} \cite{liu2018first} \\ (Freq.)\end{tabular}} & {\scriptsize $\underbrace{  \color{red}\boldsymbol{q} \boldsymbol{I} \boldsymbol{m} \boldsymbol{N} \boldsymbol{log(N)}}_{\text{T-pursuit}} + \underbrace{ I m N log(N) + I N m }_{\text{SGD in the Fourier domain}}$ }\\ \hline
		{\scriptsize SBDL} & {\scriptsize $\underbrace{ {\color{red} \boldsymbol{I} \boldsymbol{N} \boldsymbol{n} \boldsymbol{m}} + I N ( k^3 + m k^2 )}_{\text{LARS}} + \underbrace{ n m^2 }_{\text{Gram}} + \underbrace{ I N k (n + m) + n m^2 }_{\text{K-SVD}}$ }\\ \hline
        {\scriptsize Ours} & {\scriptsize $\underbrace{ {\color{red}\boldsymbol{I} \boldsymbol{N} \boldsymbol{n} \boldsymbol{m}} + I N ( k^3 + m k^2 )}_{\text{LARS}} + \underbrace{ n m^2 }_{\text{Gram}} + \underbrace{ I N (n + nk + m) }_{\text{Stochastic-LoBCoD}}$ }\\ \hline
	\end{tabular}
	\caption{Complexity analysis of our method compared to the online algorithms presented in \cite{liu2018first} and the SBDL algorithm \cite{papyan2017convolutional}. I: number of signals, N: signal dimension, m: number of filters, n: patch size, k: maximum number of non-zeros per needle, q: number of inner iterations for the pursuit algorithm.  The dominant terms are highlighted in red color.}
	\label{Table:complexity_analysis}
	\vspace{-0.5cm}
\end{table}

\noindent At this point, we turn to evaluate our proposed approach in terms of computational complexity and compare it to previous methods. We assume that the number of the non-zeros in every needle is limited to at most $k$ non-zeros, and we denote by I the number of the training images. We evaluate the complexity of every outer iteration (single epoch) of our algorithm and compare it to the complexity of executing an epoch in the alternative algorithms. Every inner iteration of our algorithm (Algorithm \ref{algorithm_2}) operates on a layer of $N/n$ needles, while the global iteration operates on the whole dataset. The resulting computation cost of the residuals $R_j$ for all the layers is $O(IN(nk+n))$, which is comprised of $O(INnk)$ for computing all the $N$ slices $\mathbf{D}_L\alpha_i$ of all the $I$ images, and $O(INn)$ computations for subtracting them from the global residual. Given the residuals, every iteration applies the LARS algorithm for solving the local sparse pursuit for all the $NI$ needles, requiring $O( k^3 + m k^2 + n m )$ per needle \cite{mairal2014sparse}, and $O( IN(k^3 + m k^2 + n m) + n m^2 )$ computations for all the $N$ needles in all the $I$ images. The latter term, $n m^2$, corresponds to the precomputation of the Gram matrix of the dictionary $\mathbf{D}_L$, which is usually negligible since it is computed once for all the needle. Next, we evaluate the complexity of reconstructing the signal. Direct computation requires $O(IN(nk+n+k))$ operations, which is effectively  $O(INnk)$, since this is the dominant term. The computation of the global residual requires another $O(IN)$ operations. These last two phases are negligible compared to the sparse pursuit stage, and are therefore omitted from the final expression. Finally, the computation of the gradient is $O(IN(n +nk))$, and the dictionary update stage is $O(INm)$ (updating the dictionary requires $O(mn)$ computations and occurs $IN/n$ times in every epoch). We summarize the above analysis in Table \ref{Table:complexity_analysis}, and compare it to the complexity analysis of the SBDL algorithm \cite{papyan2017convolutional}. In addition, Table \ref{Table:complexity_analysis} presents the complexity of executing an epoch of the two SGD based online algorithms that were introduced in \cite{liu2018first}, where q corresponds to the number of inner iterations of the sparse pursuit stage (preformed in the Fourier domain).     

The most demanding stage, in both the SBDL algorithm and in our approach, is the local sparse pursuit, which is $O(IN( k^3 + m k^2 + n m ))$. Assuming that the needles are very sparse $k\ll m$, which is often the case with real-world signals, the complexity of the local sparse pursuit stage is governed by $O(N I n m )$ in both algorithms. This implies that the complexity of our algorithm is comparable to the complexity of the SBDL, which is a batch algorithm. On the other hand, the complexity of the online algorithm in \cite{liu2018first} is dominated by the computation of the FFT in the pursuit stage, which is $O(qImNlog(N))$, meaning that their algorithm scales as $O(Nlog(N))$ with the global dimension of the signals, while our algorithm grows linearly.

\section{Image Processing via CSC}\label{sec:ImageProcessing}
Having established the foundations for our algorithms, we now set to detail their extended variants for tackling two image processing tasks: image inpainting and multi-focus image fusion.   
\subsection{Image Inpainting}\label{subsec:Inpainting}
The task of image inpainting pertains to filling-in missing pixels at known locations in the image. Assume we are given a corrupted image $Y=\mathbf{A}\signal$, where $\mathbf{A} \in \mathbb{R}^{N \times N}$ is a binary diagonal matrix that represents the degradation operator, so that $\mathbf{A}(i,i)=0$ implies that the pixel $x_i$ is masked. The goal of image inpainting is to reconstruct the original image $\signal$. Using the CSC formulation, this can be performed by first solving the following optimization problem:
\begin{equation}\label{eq:13}
\min_{\Gamma}\frac{1}{2}\|Y-\mathbf{A}\mathbf{D}\Gamma\|_2^2+\lambda\|\Gamma\|_1,
\end{equation}
and then taking the found representation $\Gamma$ and multiplying by $\mathbf{D}$. By applying the steps described in section \ref{sec:Proposed}, we split the above global optimization problem into a series of more manageable problems, each acting on a block of coordinates, i.e. a $needle$. This yields the following version of Equation (\ref{eq:9}):
\begin{equation}\label{eq:14}
\min_{\alpha_i}\frac{1}{2}\|\mathbf{P}_i R_i-\mathbf{A}_i\mathbf{D}_L\alpha_i\|_2^2+\lambda\|\alpha_i\|_1.
\end{equation}
Here, $\mathbf{A}_i=\mathbf{P}_i\mathbf{A} \mathbf{P}_i^T$ is the operator that masks the corresponding i-th patch, and $R_i =(Y-\mathbf{A}\sum_{\stackrel{j=1}{j\neq i}}^N\mathbf{P}_j^T\mathbf{D}_L\alpha_j)$ is the residual between the corrupted image and the degraded version of the reconstructed image, where the residual $R_i$ does not account for the needle $\alpha_i$. As mentioned in Section \ref{sec:Proposed}, we parallelize the computations of the needles that comprised each layer. Note that in this application, as opposed to the general case, every needle $\alpha_i$ is multiplied by a different effective dictionary $\mathbf{A}_i\mathbf{D}_L$. Consequently, the parallel computation of each needle has to also be carried out with a different effective dictionary, preventing the use of batch-LARS and other similar strategies. Yet, these pursuits are still parallelizable in different cores or nodes.

The dictionary $\mathbf{D}_L$ can be pretrained on an external, uncorrupted dataset or trained on the corrupted image directly using the following gradient:
\begin{equation}\label{eq:15} 
\nabla_{\mathbf{D}_L} =-\sum_{{i \in L_j}}\mathbf{P}_i\mathbf{A}^T(Y-\mathbf{A}\widehat{\signal})\cdotp\alpha_i^T,
\end{equation}
where $\widehat{\signal}=\sum_{j=1}^N\mathbf{P}_j^T\mathbf{D}_L\alpha_j$ is the reconstructed image. The derivation of the gradient above is identical to that described in Section \ref{sec:Dictionary Update}, with the exception of incorporating the mask $\mathbf{A}$. 

\subsection{Multi-focus image fusion} \label{subsec:Multi_focus}
Image fusion techniques aim to integrate complimentary information from multiple images, captured with different focal settings, into an all-in-focus image of higher quality. Many patch-based sparse formulations were proposed to address this task, such as choose-max OMP \cite{yang2010multifocus}, simultaneous OMP \cite{yang2012pixel}, and coupled sparse representation \cite{gao2017multi}. In this work, we adopt a similar scheme to \cite{liu2016image}, which utilizes the CSC model for tackling the task of image-fusion, with the distinction of solving a unified minimization problem.

Assume we are given a set $\{Y^k\}_{k=1}^L$ of source images to fuse, as well as a pretrained dictionary $\{d_i\}_{i=1}^m$. Each image $Y^k$ is decomposed into a base component $Y_b^k$, which is a smooth piece-wise constant image, and an edge component $Y_e^k$ that contains the high frequency elements:
\begin{equation}
Y^k = Y_b^k+Y_e^k,
\end{equation}
where the separation is performed by means of applying distinctive priors. The base component is usually extracted by imposing a prior which penalizes the $l_2$ norm of its gradient. Modeling the edge component, however, is more involved and has been the subject matter of many image-processing algorithms \cite{yang2010multifocus,yang2012pixel,gao2017multi,wang2011multi,savic2011multifocus,li1995multisensor}.
In this work, we employ the CSC model to describe the edge components, as it has shown promising results in \cite{liu2016image}.

Using the aforementioned priors, the separation of the image to its components amounts to solving the following optimization problem:
\begin{equation}\label{eq:16}
\min_{\Gamma_e^k,Y_b^k}\frac{1}{2}\|Y^k-\mathbf{D}_e\Gamma_e^k-Y_b^k\|_2^2+\lambda\|\Gamma_e^k\|_1+\mu\frac{1}{2}\|\nabla Y_b^k\|_2^2,
\end{equation}
where $\Gamma_e^k$ is the sparse representation of $Y_e^k$, under the given convolutional dictionary $\mathbf{D}_e$, i.e. $Y_e^k=\mathbf{D}_e\Gamma_e^k$, and $\|\nabla Y_b^k\|_2^2$ is given by
\begin{equation}\label{eq:17}
\|\nabla Y_b^k\|_2^2 = \|g_x\ast  Y_b^k\|_2^2+\|g_y\ast  Y_b^k\|_2^2,
\end{equation}
where $g_x = [-1~~ 1]$ and $g_y = [-1 ~~1]^T$ are the horizontal and vertical gradient operators, respectively.

By taking similar steps to those presented in section \ref{sec:Proposed}, we can once more rewrite the above optimization problem as
\begin{equation}\label{eq:18}
\begin{aligned}
\min_{\{\alpha_j^k\},Y_b^k}\frac{1}{2}\|Y^k-\sum^N_{j=1} \mathbf{P}_j^T\mathbf{D}_L\alpha_j^k-Y_b^k\|_2^2~~~~~~~\\
+\lambda\sum^N_{j=1}\|\alpha_j^k\|_1+\mu\frac{1}{2}\|\nabla Y_b^k\|_2^2,
\end{aligned}
\end{equation}
where $\{\alpha_j^k\}_{j=1}^N$ are the needles which compose the sparse vector $\Gamma_e^k$, and $\mathbf{D}_L$ is the local dictionary of $\mathbf{D}_e$.

Problem (\ref{eq:18}) can be solved by alternating between minimizing w.r.t $Y_b^k$ and $Y^k_e$, where the latter boils down to seeking for the sparse needles $\{\alpha_j^k\}_{j=1}^N$.
To that end, the update rule of the $\{\alpha_j^k\}_{j=1}^N$ is the set of local pursuit problems: 
\begin{equation}\label{eq:19}
\min_{\{\alpha_j^k\}}\frac{1}{2}\|(Y^k-Y_b^k)-\sum^N_{j=1} \mathbf{P}_j^T\mathbf{D}_L\alpha_j^k\|_2^2+\lambda\sum^N_{j=1}\|\alpha_j^k\|_1,
\end{equation}
which can be solved using our proposed algorithm,
whereas the update rule of $Y_b$ is the following least square minimization problem:
\begin{equation}\label{eq:20}
\min_{Y_b^k}\frac{1}{2}\|(Y^k-\mathbf{D}_e\Gamma_e^k)-Y_b^k\|_2^2
+\mu\frac{1}{2}\|\nabla Y_b^k\|_2^2.
\end{equation}
For solving problem (\ref{eq:20}), we set its gradient w.r.t. $Y_b$ to zero to obtain the following update rule:
\begin{equation}\label{eq:21}
Y_b^k = (I+\mu (G_x^T G_x+G_y^T G_y))^{-1}(Y^k-\mathbf{D}_e\Gamma_e^k),
\end{equation}
where $G_x$ and $G_y$ are the matrix representations of the gradient-operators.%\footnote{Due to the large size of the convolutional matrices $G_x$ and $G_y$, the matrices are not computed nor stored explicitly. Instead, this update rule can be efficiently compute using sparse matrices, or by applying the Fast Fourier Transform (FFT).}

Once these problems have been solved for all the input images $\{Y^k\}_{k=1}^L$, we aim to merge each set of feature maps\footnote{This set of feature maps refers to the $l$-th feature maps of the input images.} $\{Z_l^k\}_{k=1}^L$ in a way that best captures the focused objects in the resulting images. For each image $Y^k$, we generate an activity map based on the intensity of the $l_1$-norm of its feature maps. More specifically, we sum pixel-wise the absolute valve of its $m$ feature maps $\{Z_l^k\}_{l=1}^m$ to form an activity map that matches the size of the image $N$:
\begin{equation}\label{eq:22}
\tilde{\mathbf{A}}^k(i,j) =\sum_{l=1}^m \|Z_l^k(i,j)\|_1.
\end{equation}
To make this method more robust and less susceptible to misregistration, we convolve the above activity maps with a uniform kernel $U_s$, of a small support $s\times s$, to produce the final activity maps:
\begin{equation}\label{eq:23}
\mathbf{A}^k = \tilde{\mathbf{A}}^k\ast U_s.
\end{equation}

Based on the observation that a significant value in the activity map $\mathbf{A}^k$ indicates a sharp region in the image $Y^k$, we then reconstruct the all-in-focus edge component by selectively assembling the most prominent regions from the feature maps based on their pixel-wise values in the corresponding activity maps:
\begin{equation}
Z_l^f(i,j) = Z_l^{k^*}(i,j),~~ k^* = \argmax_k(\mathbf{A}^k (i,j)),
\end{equation}
where $\{Z_l^f\}_{l=1}^m$ are the feature maps of the fused image.
Afterward, we fuse the base components, either by taking their average
\begin{equation}
Y_b^f = \frac{1}{N}\sum_{k=1}^L Y_b^k,
\end{equation}
or by nominating regions of the base components according to the maximum value in the respective activity maps, i.e. 
\begin{equation}
Y_b^f(i,j) =Y_b^{k^*}(i,j),~~ k^* = \argmax_k(\mathbf{A}^k(i,j)),
\end{equation}
where $Y_b^f$ is the base component of the fused image.
Here, we opt for the latter since it produces better results. 

Finally, the fusion result $Y^f$ is obtained by gathering its components:
\begin{equation}
Y^f =Y_b^f+\sum_{l=1}^m d_i\ast Z_l^f.
\end{equation}

\section{Experiments}\label{sec:Experiments}
We turn to demonstrate the performance of the proposed algorithm. Throughout all our experiments we used a local dictionary composed of $m=81$ filters, each of size $8\times 8$. In addition, we used the LARS algorithm \cite{efron2004least} for solving the local sparse pursuit stage.

\subsection{Run Time Comparison}
\begin{figure}[t!]
	\centering
	\begin{subfigure}{0.23\textwidth}
		\centering
		\includegraphics[clip, trim=2.6cm 1.2cm 2.2cm 0.8cm, width=\textwidth]{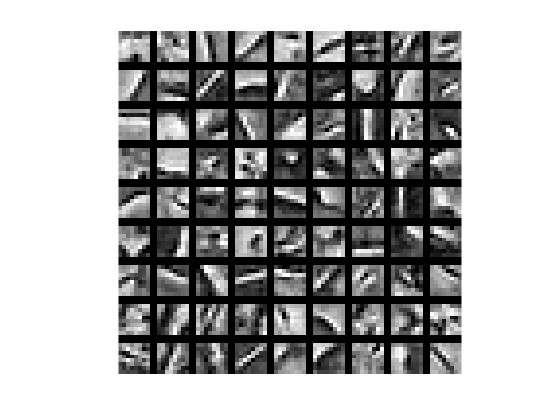}
		\caption{Ours}
	\end{subfigure}
	\begin{subfigure}{0.23\textwidth}
		\centering
		\includegraphics[clip, trim=2.6cm 1.2cm 2.2cm 0.8cm, width=\textwidth]{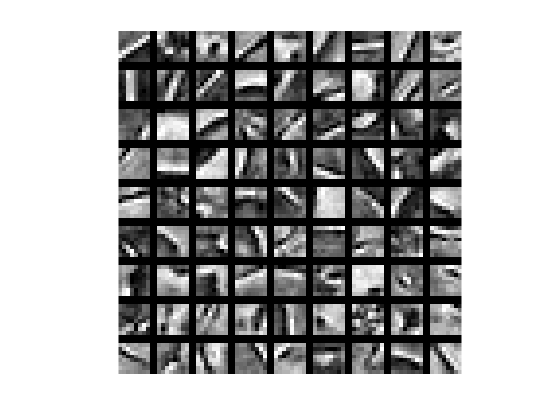}
		\caption{SBDL \cite{papyan2017convolutional} }
	\end{subfigure}
	\\[0.1cm]
	\begin{subfigure}{0.23\textwidth}
		\centering
		\includegraphics[clip, trim=2.6cm 1.2cm 2.2cm 0.8cm, width=\textwidth]{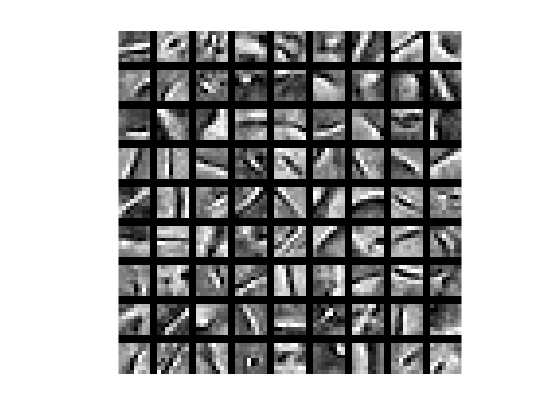}
		\caption{\cite{garcia2017subproblem}}
	\end{subfigure}
	\\[0.1cm]
	\caption{Comparison between the dictionaries obtained using the Stochastic-LoBCoD method vs. the methods in \cite{papyan2017convolutional} and \cite{garcia2017subproblem} on the Fruit dataset. }
	\label{figure:Dictionaries_batch}
\end{figure}

To begin with, and to provide a comparison to other state of the art methods, we evaluate the performance of the proposed algorithm for solving Equation (\ref{eq:4}) against other leading batch algorithms for CSC: the SBDL algorithm \cite{papyan2017convolutional}, the algorithm in \cite{wohlberg2016boundary} and the algorithm presented in \cite{garcia2017subproblem}, all ran with $\lambda=20$ on the Fruit dataset \cite{zeiler2010deconvolutional}. The dataset contains 10 images of size $100\times 100$ pixels, and all the images were mean-subtracted by convolving them with an $8 \times 8$ uniform kernel as a preprocessing step. For learning the dictionary, we used the ADAM algorithm  \cite{ruder2016overview} in the initial 30 iterations, with $\eta =0.02$, and instate the ADAM parameters in accordance with the authors' recommendation: $\beta_1=0.9$, $\beta_2=0.999$, and $\epsilon=10^{−8}$. Subsequent iterations applied the Momentum algorithm with $\eta=10^{-7}$ and $\gamma =0.8$ until convergence\footnote{All our notations are in accordance with those presented in \cite{ruder2016overview}}. Fig. \ref{figure:Objective_Vs_time_compar} presents a comparison of the objective value as a function of time for each of the competing algorithms, showing that our method achieves the fastest convergence. Fig. \ref{figure:Dictionaries_batch} shows the dictionaries obtained by our method, and the batch methods in \cite{papyan2017convolutional} and \cite{garcia2017subproblem}. Note that the obtained dictionaries tend to look similar.

\begin{figure}[t!]
	\begin{subfigure}{0.23\textwidth}
		\centering
		\includegraphics[clip, trim=2.6cm 1.2cm 2.2cm 0.8cm, width=\textwidth]{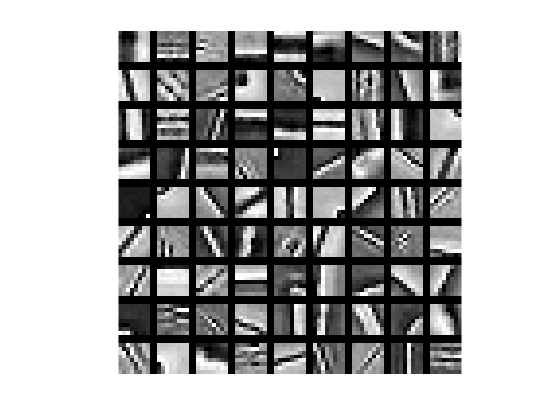}
		\caption{Ours}
	\end{subfigure}
	\begin{subfigure}{0.23\textwidth}
		\centering
		\includegraphics[clip, trim=2.6cm 1.2cm 2.2cm 0.8cm, width=\textwidth]{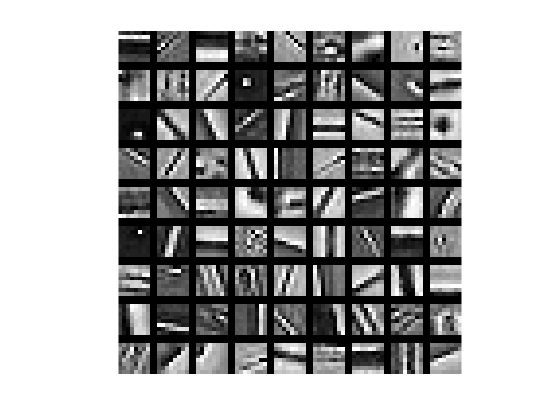}
		\caption{\cite{liu2018first} Sparse}
	\end{subfigure}
	\\[0.1cm]
	
	~~ ~~ ~~ ~~ ~~ ~~       
	\begin{subfigure}{0.23\textwidth}
	    \centering
		\includegraphics[clip, trim=2.5cm 1.2cm 2.2cm 0.8cm, width=\textwidth]{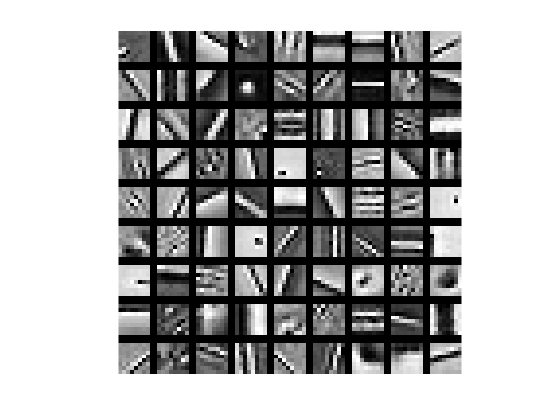}
		\caption{\cite{liu2018first} Frequency}
	\end{subfigure}
	\\[0.1cm]
	\caption{Comparison between the dictionaries obtained using the Stochastic-LoBCoD method vs. the online methods in \cite{liu2018first} in the original and frequency domains, all run on the MIRFLICKR-1M dataset.}
	\label{figure:Dictionaries_online}
\end{figure}

We also compared our method to the online, stochastic gradient descent (SGD) based algorithms in \cite{liu2018first}, which operate in the spatial and in the Fourier domains. In this comparison we used 40 images randomly selected from the MIRFLICKR-1M dataset \cite{huiskes2010new} for training the dictionaries, as well as a test set of 5 different images from the same source. The images were cropped to reduce their size from $512\times512$ to $256\times256$ pixels in both the training and testing sets to expedite the computation. In addition, we divided the images by 255 and mean-subtracted them as was done in the Fruit dataset. In this experiment, we used $\lambda=0.1$ and a learning rate of $\eta =0.1$, with learning rate decay of $\eta/(1+200/t)$ every 5 epochs, and momentum with $\gamma =0.8$. Fig. \ref{figure:Objective_Vs_time_Online} presents the objective of the test set as a function of time, showing that our algorithm converges faster. Fig \ref{figure:Dictionaries_online} shows the dictionaries obtained by the three method, illustrating similar quality. 
%In addition, the proposed method obtains a slightly smaller minimum objective value than the other five methods, which becomes increasingly apparent as the number of iterations grows.
%Note, however, that the online algorithms in \cite{liu2018first} are tailored for large training sets, rather than the relatively small one that we study in this work. Therefore, these algorithms show poor performance for this setting as expected.

\textbf{\begin{figure}[t] 
    \centering
     \includegraphics[clip, trim=3.7cm 8.8cm 3.8cm 9.5cm, width=0.5\textwidth]{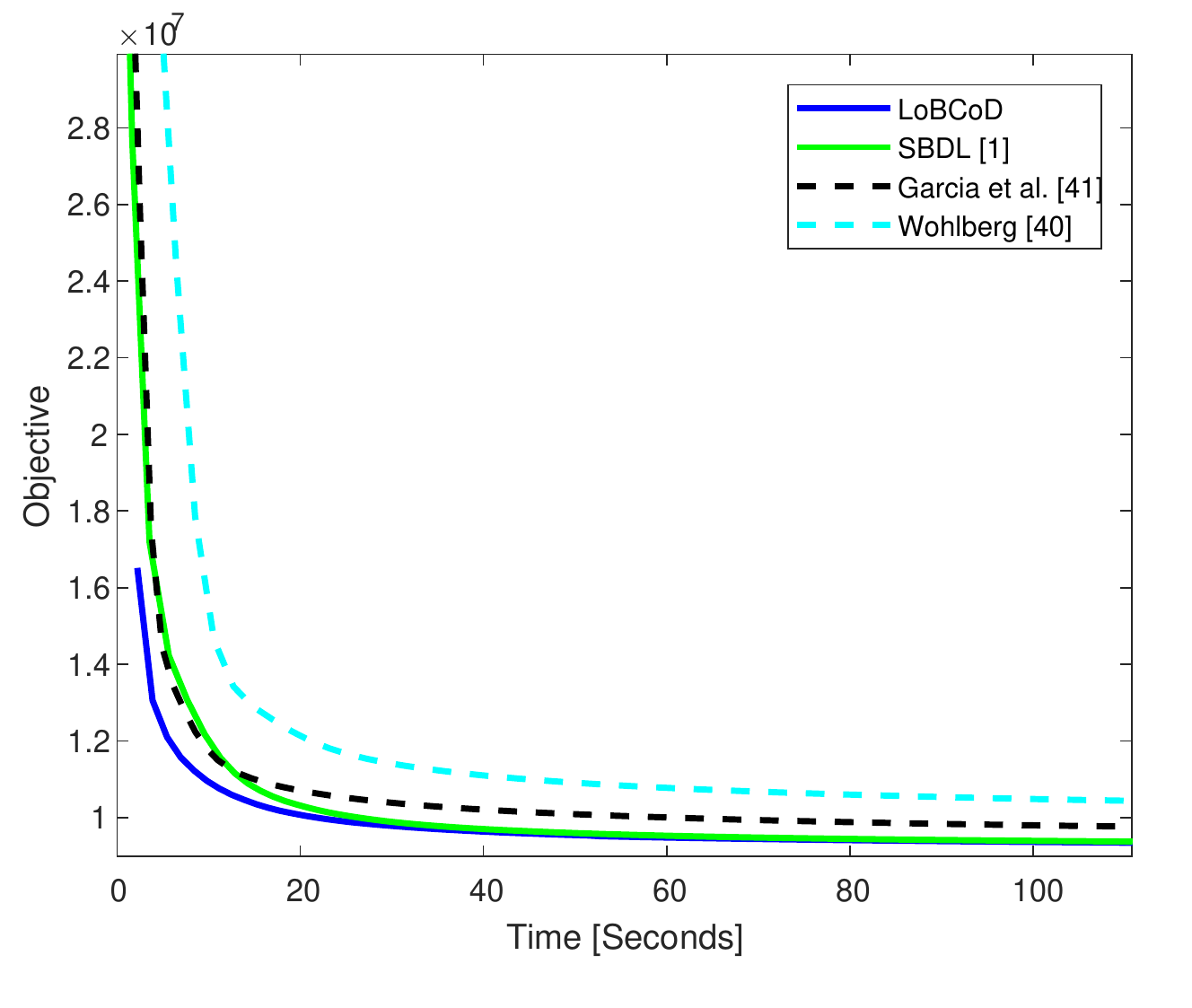}
     \caption{Run time comparison between our method and the batch methods: the SBDL algorithm \cite{papyan2017convolutional}, the algorithm of Wohlberg \cite{wohlberg2016boundary} and the algorithm by Garcia et al. \cite{garcia2017subproblem}.}  
     \label{figure:Objective_Vs_time_compar}
\end{figure}}
%[width=0.37\textwidth]

\textbf{\begin{figure}[t!] 
    \centering
     \includegraphics[clip, trim=3.7cm 8.8cm 3.8cm 9.5cm, width=0.5\textwidth]{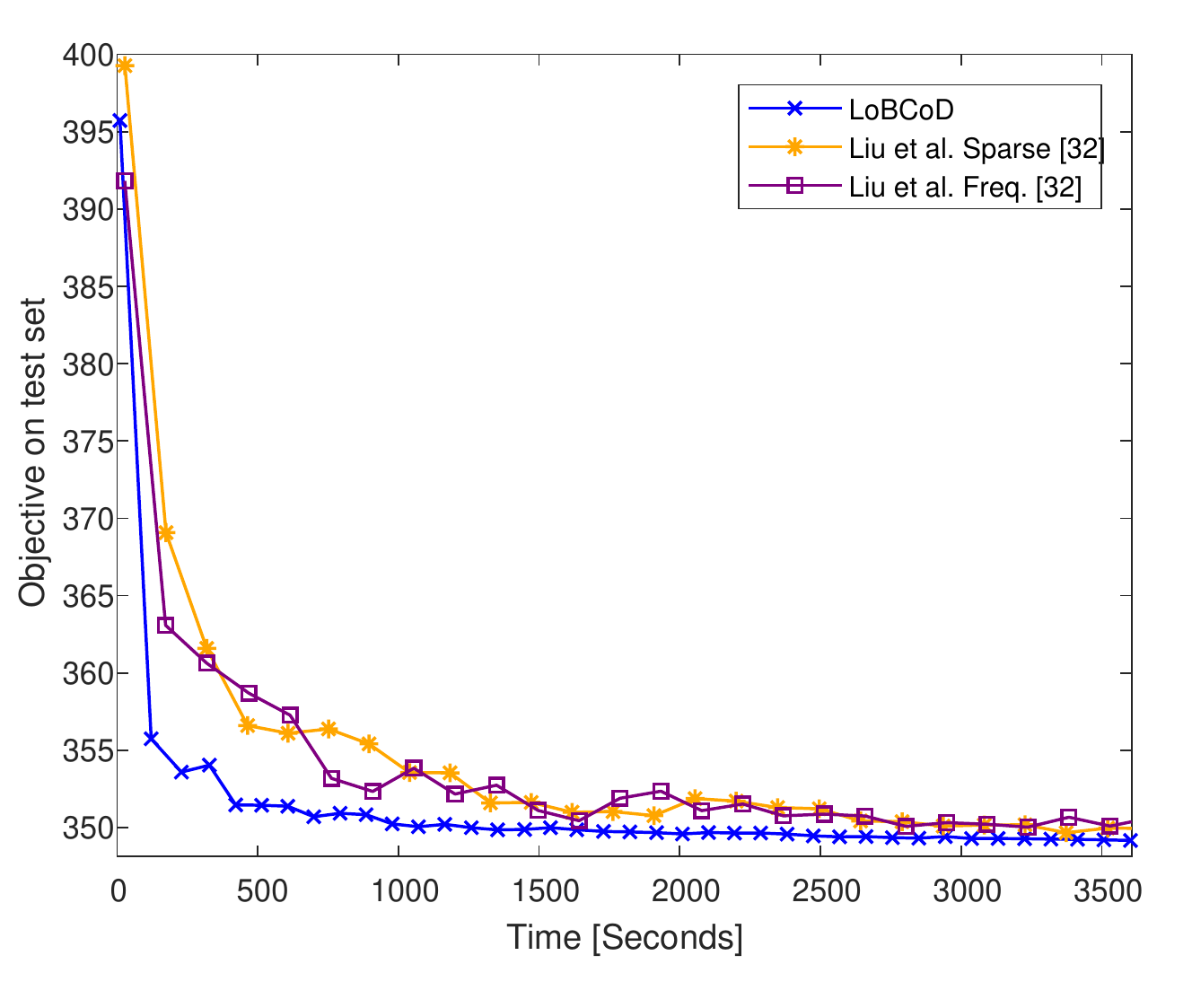}
     \caption{Run time comparison between our method and the online algorithms of Liu at el. \cite{liu2018first} in the original and frequency domains.}  
     \label{figure:Objective_Vs_time_Online}
\end{figure}}

% \begin{figure}[b!]
% 	\centering
% 	\begin{subfigure}{0.235\textwidth}
% 		\centering
% 		\includegraphics[clip, trim=1.2cm 1.2cm 1.2cm 0.8cm, width=\textwidth]{Barbara_BCD_dictionary.png}
% 	\end{subfigure}
% 	\begin{subfigure}{0.235\textwidth}
% 		\centering
% 		\includegraphics[clip, trim=1.2cm 1.2cm 1.2cm 0.8cm, width=\textwidth]{Barbara_SBDL_dictionary.png}
% 	\end{subfigure}
% 	\caption{The dictionaries obtained by training on the corrupted image \textsf{\small{Barbara}}. Left: Ours (PSNR = 32.5dB). Right: The SBDL \cite{papyan2017convolutional} (PSNR = 31.98dB).}
% 	\label{Figure:Dictionaries_Barbara}
% \end{figure}

\begin{table*}[ht]
\caption{Inpainting comparison between the proposed Stochastic-LoBCoD and the SBDL \cite{papyan2017convolutional} algorithms.}
\label{table_inpainting}
\begin{center}
\begin{tabular}{|c||c||c||c||c||c||c||c||c||c||c||c|}
\hline
  & Barbara & Boat & House & Lena & Peppers & C.man & Couple & Finger & Hill & Man & Montage\\
\hline
\hline
SBDL (external)& 30.41 &  31.76 &  36.17 &  35.92 & \textbf{33.69}  & 28.76  & 32.16  &  30.91 & 33.12  &  33.04 & 28.93 \\
\hline
Proposed (external)& \textbf{30.93} &  \textbf{31.82} &  \textbf{36.58} & \textbf{36.15}  &  33.54 & \textbf{28.88}  & \textbf{32.46}  &  \textbf{31.75} &  \textbf{33.25} &  \textbf{33.18} &  \textbf{29.18}\\
\hline
\hline
Image specific SBDL & 31.98 & 32.04  & 36.19 &  36.01 & 34.03 &  28.85 &  32.18 & 30.96  & 33.21  &  32.99 & 28.95 \\
\hline
Image specific Proposed & \textbf{32.50}  & \textbf{32.27}  &  \textbf{36.74} &  \textbf{36.17} &  \textbf{34.48} & \textbf{29.04} & \textbf{32.56}  &  \textbf{31.76} & \textbf{33.42}  & \textbf{33.25} & \textbf{29.23} \\ 
\hline
\end{tabular}
\end{center}
\end{table*}

\begin{figure*}[ht]
	\centering
	\begin{subfigure}{0.235\textwidth}
		\centering
		\includegraphics[clip, trim=1.8cm 1.0cm 1.0cm 0cm, width=\textwidth]{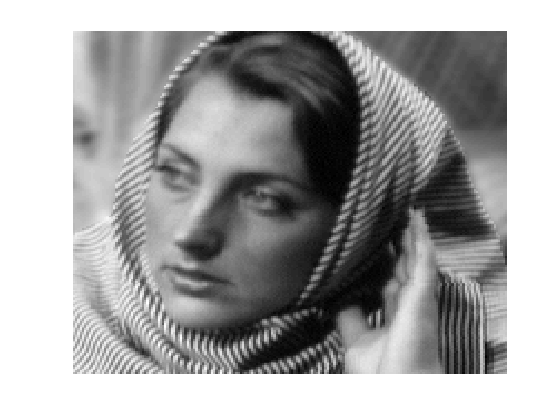}
        \caption{\textsf{\small{Barbara}} in-focus}\label{Figure:Barbara_back}
	\end{subfigure}
	\begin{subfigure}{0.235\textwidth}
		\centering
		\includegraphics[clip, trim=1.8cm 1.0cm 1.0cm 0cm, width=\textwidth]{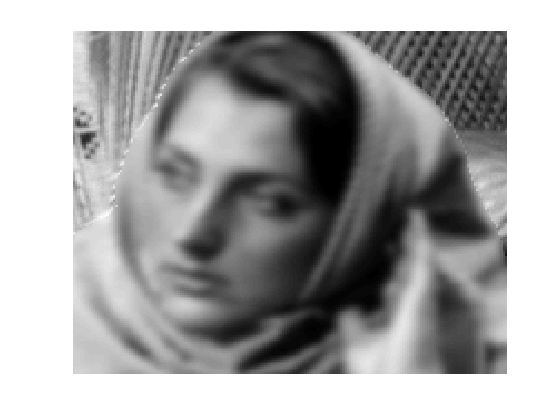}
        \caption{\textsf{\small{Barbara}} out-of-focus}
	\end{subfigure}
    \begin{subfigure}{0.235\textwidth}
		\centering
		\includegraphics[clip, trim=1.8cm 1.0cm 1.0cm 0cm, width=\textwidth]{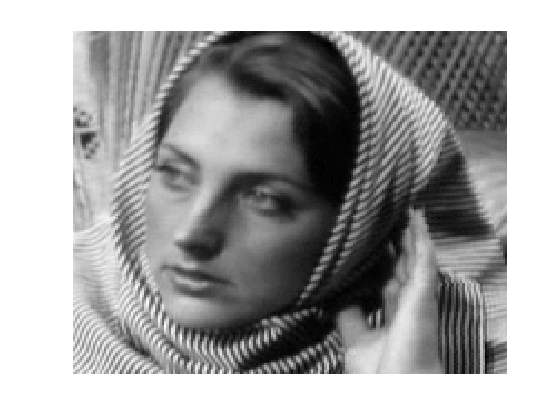}
        \caption{\cite{liu2016image} PSNR: 41.96dB }
	\end{subfigure}
    \begin{subfigure}{0.235\textwidth}
		\centering
		\includegraphics[clip, trim=1.8cm 1.0cm 1.0cm 0cm, width=\textwidth]{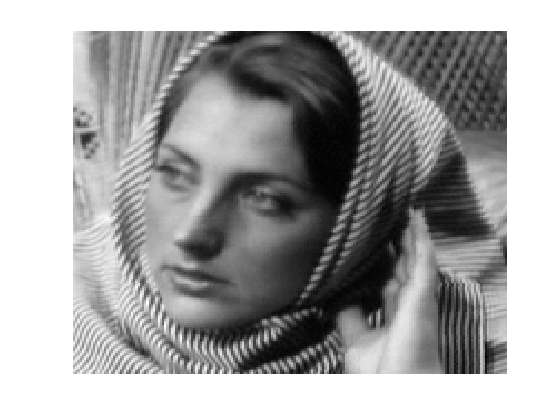}
        \caption{Proposed PSNR: 42.27dB }
	\end{subfigure}
    	\begin{subfigure}{0.235\textwidth}
		\centering
		\includegraphics[clip, trim=1.8cm 1.0cm 1.0cm 0cm, width=\textwidth]{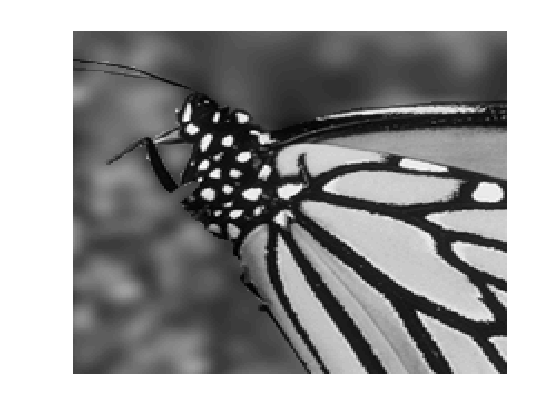}
        \caption{\textsf{\small{Butterfly}} in-focus}
	\end{subfigure}
	\begin{subfigure}{0.235\textwidth}
		\centering
		\includegraphics[clip, trim=1.8cm 1.0cm 1.0cm 0cm, width=\textwidth]{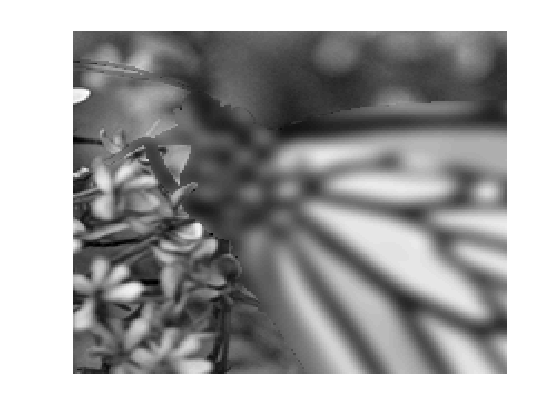}
        \caption{\textsf{\small{Butterfly}} out-of-focus}
	\end{subfigure}
    \begin{subfigure}{0.235\textwidth}
		\centering
		\includegraphics[clip, trim=1.8cm 1.0cm 1.0cm 0cm, width=\textwidth]{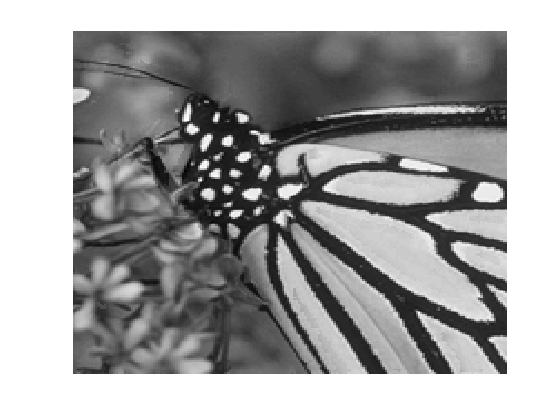}
        \caption{\cite{liu2016image} PSNR: 35.30dB }
	\end{subfigure}
    \begin{subfigure}{0.235\textwidth}
		\centering
		\includegraphics[clip, trim=1.8cm 1.0cm 1.0cm 0cm, width=\textwidth]{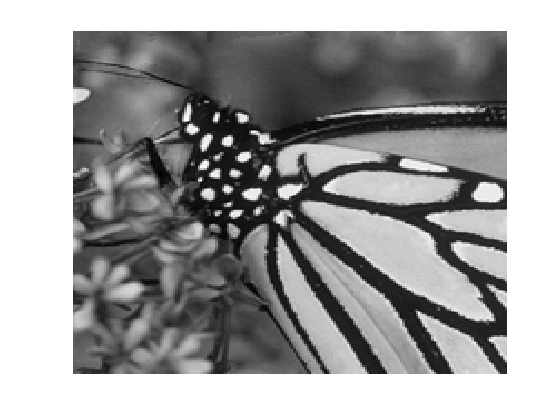}
        \caption{Proposed PSNR: 36.01dB }
	\end{subfigure}
    %\\[0.5cm]
	\caption{Fusion performance comparison between our algorithm and the results of \cite{liu2016image} on synthetic images. The PSNR values were computed between the reconstructed and the original images.\\ \\}
	\label{Figure:Syntetic_Images}
\end{figure*}

\begin{figure*}[ht]
	\centering
	\begin{subfigure}{0.235\textwidth}
		\centering
		\includegraphics[clip, trim=1.8cm 1.0cm 1.0cm 0cm, width=\textwidth]{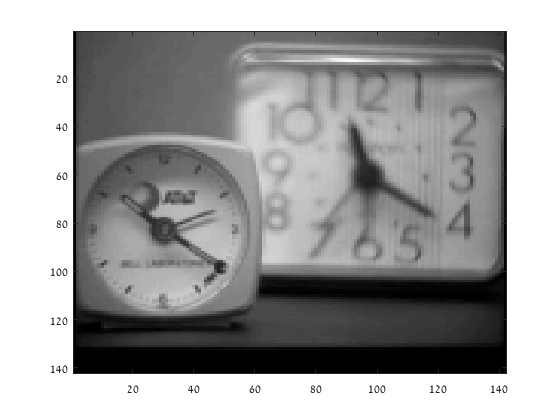}
        \caption{Near in-focus}
	\end{subfigure}
	\begin{subfigure}{0.235\textwidth}
		\centering
		\includegraphics[clip, trim=1.8cm 1.0cm 1.0cm 0cm, width=\textwidth]{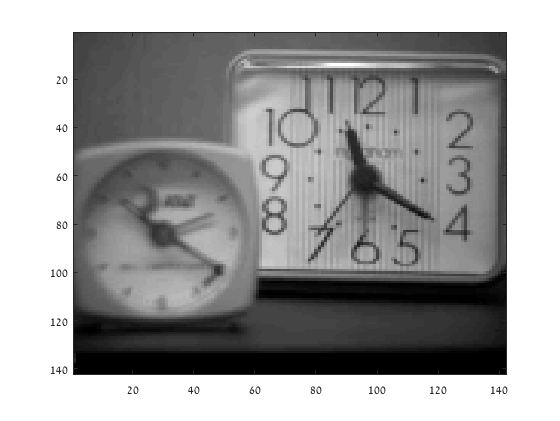}
        \caption{Far in-focus}
	\end{subfigure}
    \begin{subfigure}{0.235\textwidth}
		\centering
		\includegraphics[clip, trim=1.8cm 1.0cm 1.0cm 0cm, width=\textwidth]{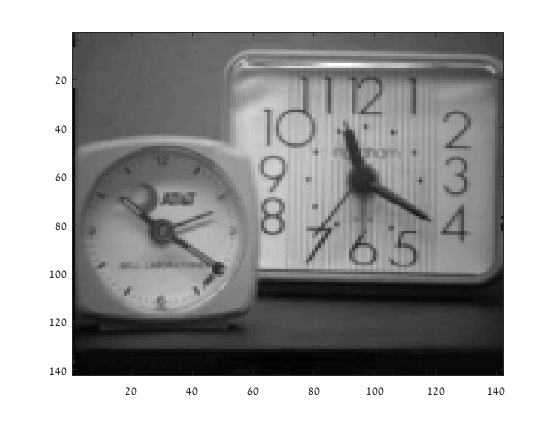}
        \caption{\cite{liu2016image}}
	\end{subfigure}
    \begin{subfigure}{0.235\textwidth}
		\centering
		\includegraphics[clip, trim=1.8cm 1.0cm 1.0cm 0cm, width=\textwidth]{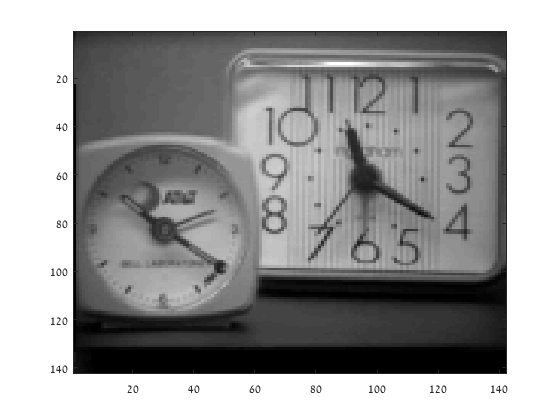}
        \caption{Proposed}
	\end{subfigure}
    	\begin{subfigure}{0.236\textwidth}
		\centering
		\includegraphics[clip, trim=1.8cm 1.9cm 1.0cm 0.0cm, width=\textwidth]{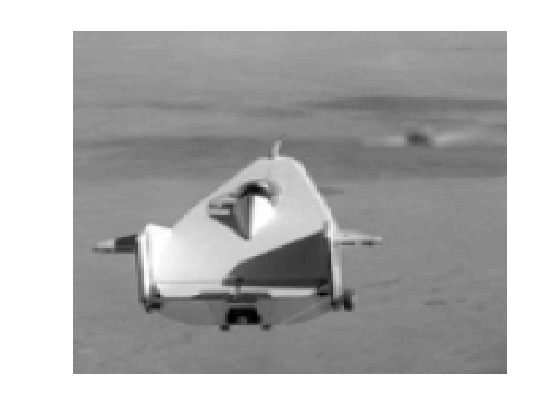}
        \caption{Near in-focus}
	\end{subfigure}
	\begin{subfigure}{0.236\textwidth}
		\centering
		\includegraphics[clip, trim=1.8cm 1.9cm 1.0cm 0.0cm, width=\textwidth]{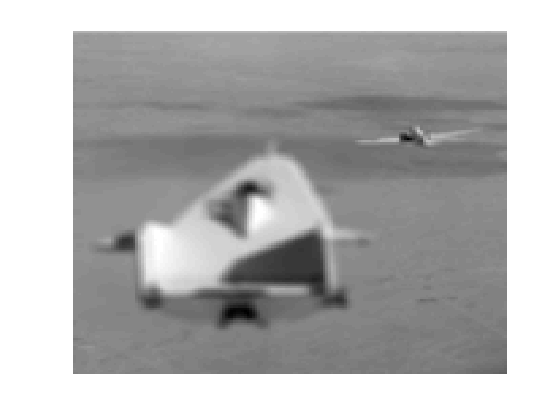}
        \caption{Far in-focus}
	\end{subfigure}
    \begin{subfigure}{0.236\textwidth}
		\centering
		\includegraphics[clip, trim=1.8cm 1.9cm 1.0cm 0.0cm, width=\textwidth]{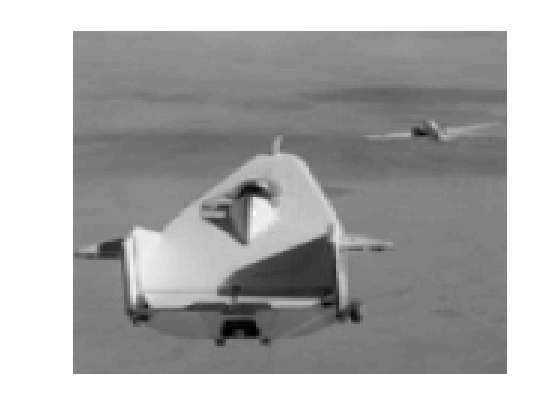}
        \caption{\cite{liu2016image}}
	\end{subfigure}
    \begin{subfigure}{0.236\textwidth}
		\centering
		\includegraphics[clip, trim=1.8cm 1.9cm 1.0cm 0.0cm, width=\textwidth]{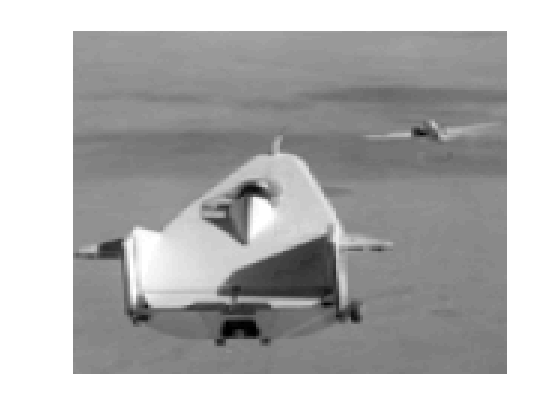}
        \caption{Proposed}
	\end{subfigure}
    %\\[0.5cm]
	\caption{Fusion examples of real images taken with different focal settings. The images were taken from the dataset in \cite{savic2011multifocus}.\\ }
	\label{Figure:Real_Images}
\end{figure*}

\begin{figure}[t!]
	\centering
	\begin{subfigure}{0.15\textwidth}
		\centering
		\includegraphics[clip, trim=2cm 1.0cm 1.5cm 0cm, width=\textwidth]{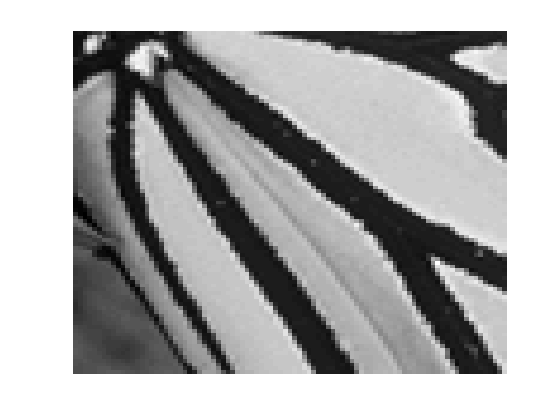}
		\caption{Original}
	\end{subfigure}
	\begin{subfigure}{0.15\textwidth}
		\centering
		\includegraphics[clip, trim=2cm 1.0cm 1.5cm 0cm, width=\textwidth]{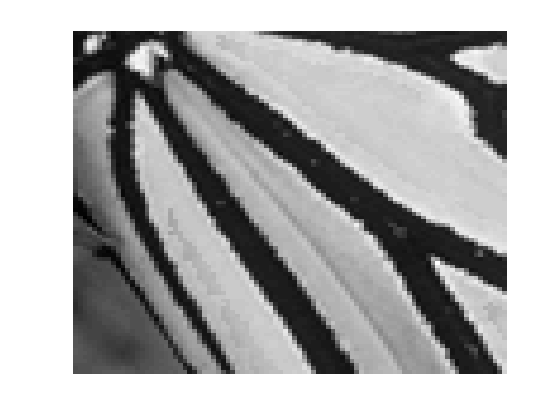}
		\caption{\cite{liu2016image} }
	\end{subfigure}
		\begin{subfigure}{0.15\textwidth}
		\centering
		\includegraphics[clip, trim=2cm 1.0cm 1.5cm 0cm, width=\textwidth]{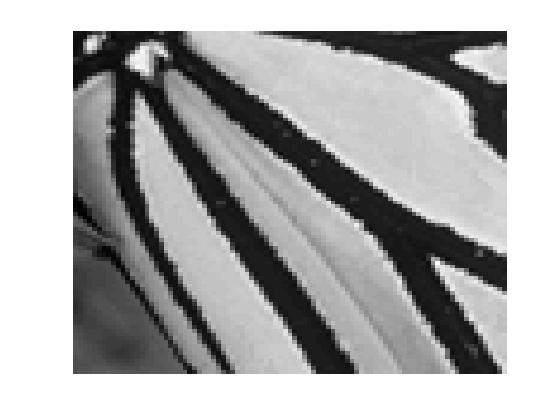}
		\caption{Proposed}
	\end{subfigure}
		\begin{subfigure}{0.15\textwidth}
		\centering
		\includegraphics[clip, trim=2cm 1.0cm 1.5cm 0cm, width=\textwidth]{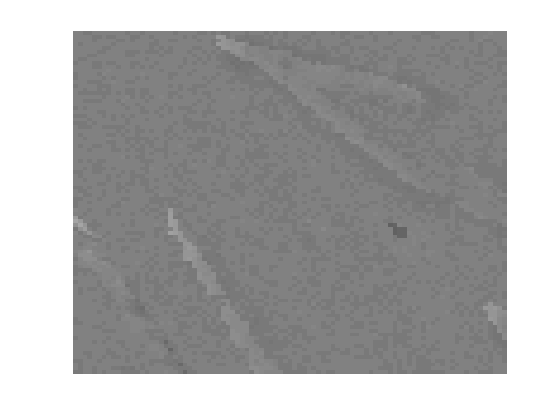}
		\caption{\cite{liu2016image}}
	\end{subfigure}
		\begin{subfigure}{0.15\textwidth}
		\centering
		\includegraphics[clip, trim=2cm 1.0cm 1.5cm 0cm, width=\textwidth]{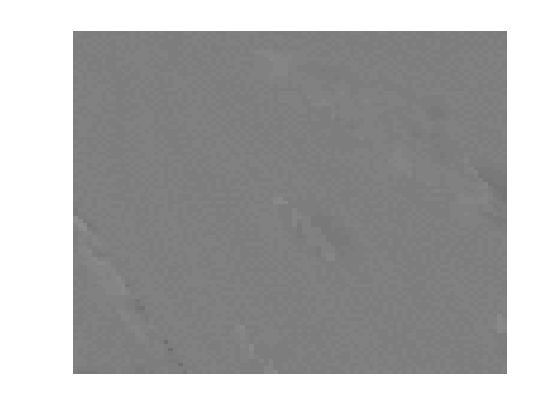}
		\caption{Proposed}
	\end{subfigure}
	\caption{Zoom in on the fusion results of the image \textsf{\small{Butterfly}}. Figures (d) and (e) present the error images of (b) and (c), respectively. The error images were computed between the fusion results and the original image (a).}
	\label{Fig:zoomin}
	%\vspace{-0.5cm}
\end{figure}

\subsection{Image Inpainting}

We applied our algorithm to the task of image inpainting, as described in Section \ref{subsec:Inpainting}, and compared our results to \cite{papyan2017convolutional}, which was shown to provide the best performance in this task amongst all previous methods. In Section \ref{subsec:Inpainting} we described two viable methods for training the dictionary; the first approach is to utilize an external dataset, while the second is to train directly on the corrupted source image itself.
For training the dictionary using an external dataset, we used the Fruit dataset \cite{zeiler2010deconvolutional} for both algorithms, as shown in Fig. \ref{figure:Dictionaries_batch}. All the corrupted test images were mean-subtracted prior to applying both algorithms by computing the patch-average of each pixel using only the unmasked pixels, and subtracting the resulting mean image from the original one. In addition, we tuned $\lambda$ in Equation (\ref{eq:13}) for every corrupted test image, to account for their varying complexity. The top two rows of Table \ref{table_inpainting} present the results using an external dataset in terms of peak signal-to-noise ratio (PSNR)\footnote{The PSNR is computed as $20log(255\sqrt{N}/\|\signal-\widehat{\signal}\|_2)$, where $\signal$ and $\widehat{\signal}$ are the original and the restored images. Note that in contrast to \cite{papyan2017convolutional}, here the images were only mean subtracted, thus their original gray scale range is preserved.} on a set of 11 standard test images, showing that our method leads to quantitatively better results. Next, we train the dictionary of both algorithms on the corrupted image itself, where in our method we use the update rule for the gradient as described in Equation (\ref{eq:15}). The results are presented in the last two rows of Table \ref{table_inpainting}, indicating that the proposed Stochastic-LoBCoD algorithm achieves quantitatively better results.

\begin{figure}[ht]
	\centering
	\begin{subfigure}{0.239\textwidth}
		\centering
		\includegraphics[clip, trim=2cm 1.0cm 1.5cm 0cm, width=\textwidth]{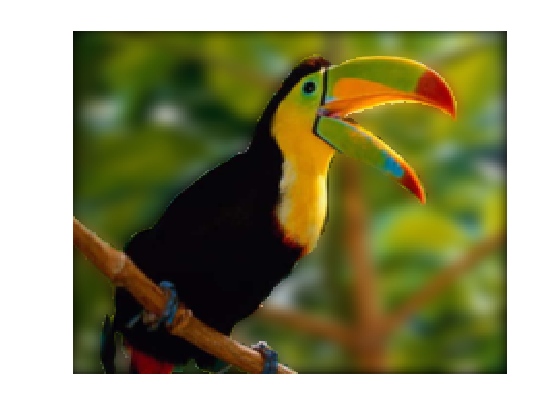}
		\caption{\textsf{\small{Bird}} Foreground}
	\end{subfigure}
	\begin{subfigure}{0.239\textwidth}
		\centering
		\includegraphics[clip, trim=2cm 1.0cm 1.5cm 0cm, width=\textwidth]{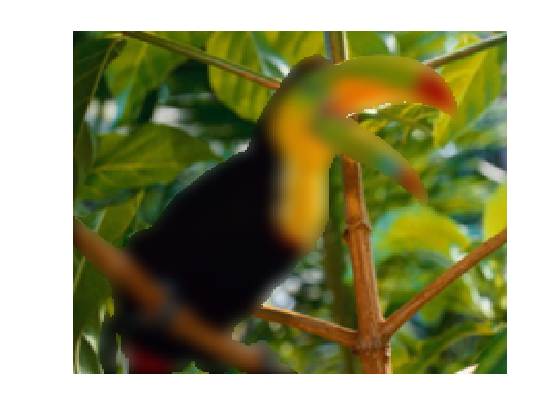}
		\caption{\textsf{\small{Bird}} Background}
	\end{subfigure}
		\begin{subfigure}{0.239\textwidth}
		\centering
		\includegraphics[clip, trim=2cm 1.0cm 1.5cm 0cm, width=\textwidth]{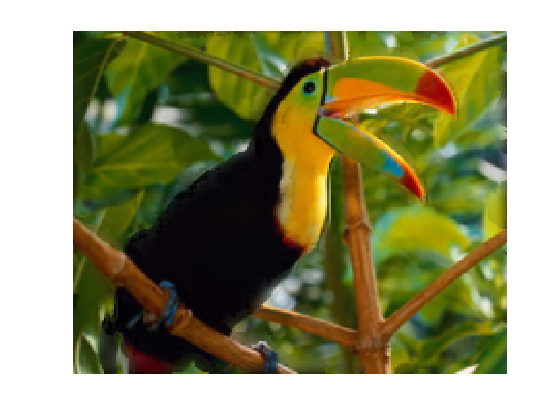}
		\caption{\cite{liu2016image} PSNR: 39.29dB}
	\end{subfigure}
		\begin{subfigure}{0.239\textwidth}
		\centering
		\includegraphics[clip, trim=2cm 1.0cm 1.5cm 0cm, width=\textwidth]{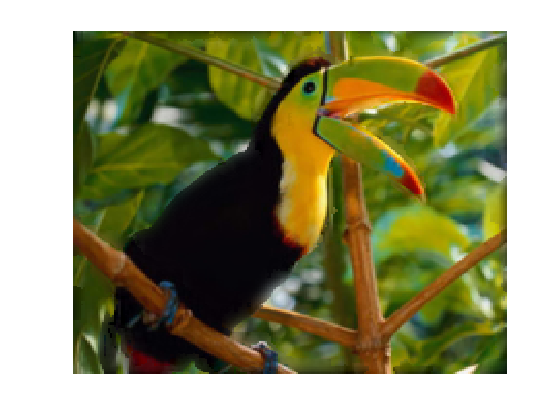}
		\caption{Proposed PSNR: 39.81dB}
	\end{subfigure}
	\caption{Fusion comparison between the proposed method and the method in [21] on the image \textsf{\small{Bird}}.}
	\label{Fig:Bird}
	\vspace{-0.5cm}
\end{figure}

\begin{figure}[t]
	\centering
	\begin{subfigure}{0.11\textwidth}
		\centering
		\includegraphics[clip, trim=6cm 6cm 6cm 2cm, width=\textwidth]{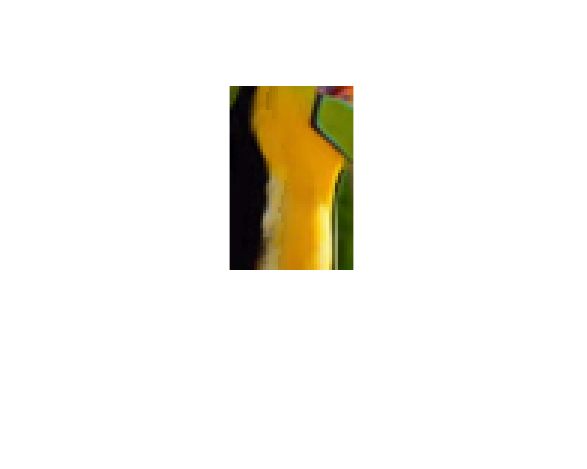}
		\caption{\cite{liu2016image}}
	\end{subfigure}
	    \begin{subfigure}{0.11\textwidth}
		\centering
		\includegraphics[clip, trim=6cm 6cm 6cm 2cm, width=\textwidth]{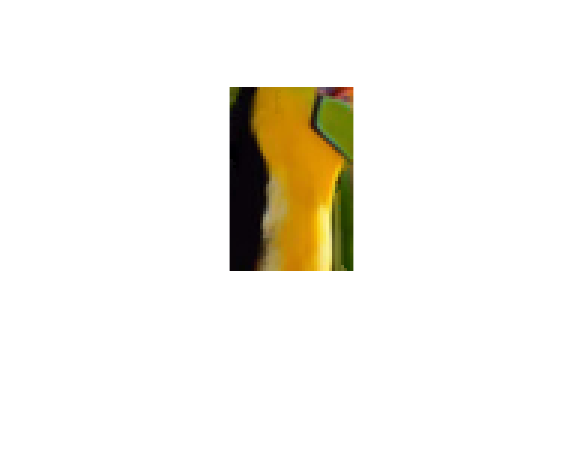}
		\caption{Proposed}
	\end{subfigure}
		\begin{subfigure}{0.11\textwidth}
		\centering
		\includegraphics[clip, trim=6cm 6cm 6cm 2cm, width=\textwidth]{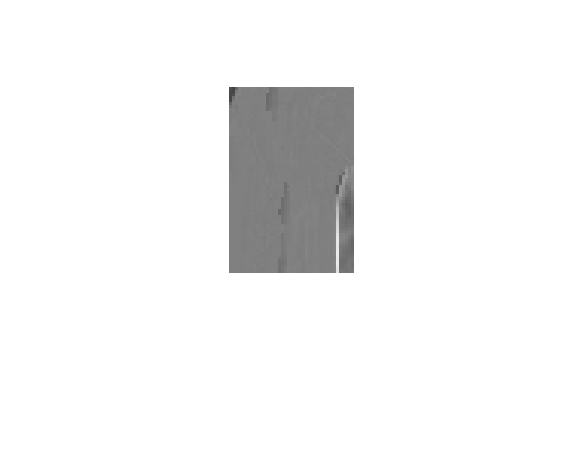}
		\caption{\cite{liu2016image}}
	\end{subfigure}
		\begin{subfigure}{0.11\textwidth}
		\centering
		\includegraphics[clip, trim=6cm 6cm 6cm 2cm, width=\textwidth]{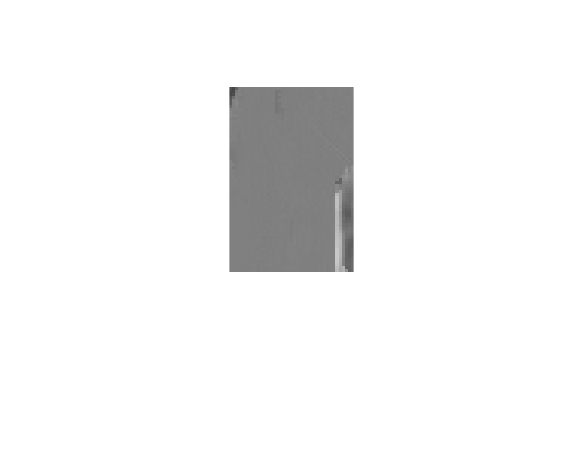}
		\caption{Proposed}
	\end{subfigure}
	\caption{Zoom in on the fusion results of the image \textsf{\small{Bird}}. Figure (c) and (d) present the L channel error compared to that of the original image.}
	\label{Fig:Bird_ZoomIn}
	\vspace{-0.5cm}
\end{figure}

\subsection{Multi-focus image fusion} 
We conclude by applying our LoBCoD algorithm to the task of multi-focus image fusion, as described in Section \ref{subsec:Multi_focus}. We evaluate our proposed method using synthetic data, as well as data from a real dataset, and compare our results to the one reported in \cite{liu2016image}. The dictionaries of both methods were pretrained on the Fruit dataset \cite{zeiler2010deconvolutional}. In addition, we set the coefficients described in Equation (\ref{eq:18}) to $\lambda =1$ and $\mu=5$ for the sparse pursuit and the base-image extraction stages, respectively, and alternate between the stages at each iteration. In practice, convergence was achieved within 2-4 iterations of alternating between these two stages.

For the synthetic experiment, we extracted a portion of the standard image \textsf{\small{Barbara}} and created two input images, one with a blurred foreground and another with a blurred background. Image blurring was performed using a $9 \times 9$ Gaussian blur kernel with $\sigma=2$. The size of the reconstruction kernel $U_s$, presented in Equation (\ref{eq:23}) was chosen to be $9 \times 9$. We repeated the same procedure on the image \textsf{\small{Butterfly}}\footnote{The image was taken form the dataset in \cite{dong2013nonlocally}.}, using a $16 \times 16$ Gaussian blur kernel with $\sigma=4$, and a reconstruction kernel $U_s$ of size $8 \times 8$. Both sets of synthetic blurred images are presented in Fig. \ref{Figure:Syntetic_Images}, alongside their reconstructed images. The PSNR values between the reconstructed images and the original ones are also detailed in Fig. \ref{Figure:Syntetic_Images}. The resulting images demonstrate that our approach leads to visually and quantitatively better results. Fig. \ref{Fig:zoomin} presents a zoom-in view of our reconstructed image \textsf{\small{Butterfly}}, compared to the result of \cite{liu2016image} and the original image; showing that for images with prominent blur, as in the case of the image \textsf{\small{Butterfly}}, our method achieves visually better results.

We adapt our approach for fusion of colored images. We blurred the image \textsf{\small{Bird}}\footnote{The image was taken form the dataset in \url{https://github.com/titu1994/Image-Super-Resolution/tree/master/val_images/set5}} by applying a $16 \times 16$ Gaussian blur kernel with $\sigma=4$ on each channel of the RGB color space separately, to create the foreground and the background blurred images. We chose to blur the image in the RGB color space to emulate a blur of a camera. Afterwords, both blurred colored images were treated by transforming them to the Lab color space, and building the activity maps $A^k$ based on their L channel, with kernel $U_s$ of size of  $14\times 14$. Then, we reconstructed each channel from the Lab color space by selecting regions based on the maximum pixel-wise value of the activity maps. The PSNR for the \textsf{\small{Bird}} image was computed between the L channels of the original and the reconstructed images. We present the results together with their PSNR values in Fig. \ref{Fig:Bird}, which shows that our approach leads to visually and quantitatively better results.

Lastly, for the experiment with the real dataset, we ran our proposed algorithm on two sets of images: the \textsf{\small{Clocks}} and the \textsf{\small{Planes}}, both taken from the dataset in \cite{savic2011multifocus}. To fuse the \textsf{\small{Clocks}} images we used a uniform kernel of size $9 \times 9$, whereas for the \textsf{\small{Planes}} images we used a $16 \times 16$ uniform kernel. Fig. \ref{Figure:Real_Images} presents the resulting fused images, showing comparable results for both algorithms on this dataset.

%Lastly, we adapt our approach for fusion of colored images. We blurred the image \textsf{\small{Bird}}\footnote{The image was taken form the dataset in \url{https://github.com/titu1994/Image-Super-Resolution/tree/master/val_images/set5}} twice by applying a $16 \times 16$ Gaussian blur kernel with $\sigma=4$ on each channel of the RGB color space separately, to create the foreground and the background blurred images. We chose to blur the image in the RGB color space to emulate a blur of a camera. Afterwords, both blurred colored images were treated by transforming them to the Lab color space, and building the activity maps $A^k$ based on their L channel, with kernel $U_s$ of size of  $14\times 14$. Then, we reconstructed each channel from the Lab color space by selecting regions based on the maximum pixel-wise value of the activity maps. The PSNR for the \textsf{\small{Bird}} image was computed between the L channels of the original and the reconstructed images. We present the results together with their PSNR values in Fig. \ref{Fig:Bird}, which shows that our approach leads to visually and quantitatively better results.

\section{Conclusions}\label{sec:Conclusions}
In this work we have introduced the local block coordinate descent (LoBCoD) algorithm for performing pursuit for the global CSC model, while operating locally on image patches. We demonstrated its advantages over contending state-of-the-art methods in terms of memory requirements, efficient parallel computation, and its exemption from meticulous manual tuning of parameters. In addition, we proposed a stochastic gradient descent version (Stochastic-LoBCoD) of this algorithm for training the convolutional filters. We highlighted its unique qualities as an online algorithm that retains the ability to act on a single image. %; leveraging the benefits of online learning, while learning filters that reliably describe the signal at hand. 
Finally, we illustrated the advantages of the proposed algorithm on a set of applications and compared it with competing state-of-the-art methods.

% A prospective future direction would be an extension of the LSGD algorithm to multi-layer CSC setting. This architecture would allow training a complex multi-layer global model on a single image. This way, we will benefit from the global properties of the representation and the stochastic advantages in the setting of working on non-convex optimization, while learning relevant filters for the signal at hand.         


%%%%%%%%%%%%%%%%%%%%%%%%%%%%%%%%%%%%%%%%%%%%%%%%%%%%%%%%%%%%%%%%%%%%%%%%%%%%%%%%

%%%%%%%%%%%%%%%%%%%%%%%%%%%%%%%%%%%%%%%%%%%%%%%%%%%%%%%%%%%%%%%%%%%%%%%%%%%%%%%%

%%%%%%%%%%%%%%%%%%%%%%%%%%%%%%%%%%%%%%%%%%%%%%%%%%%%%%%%%%%%%%%%%%%%%%%%%%%%%%%%

\section*{Acknowledgment}

The research leading to these results has received funding in part from the European Research Council under EU’s 7th Framework Program, ERC under Grant 320649, and in part by Israel Science Foundation (ISF) grant no. 335/18.

%%%%%%%%%%%%%%%%%%%%%%%%%%%%%%%%%%%%%%%%%%%%%%%%%%%%%%%%%%%%%%%%%%%%%%%%%%%%%%%%

\bibliographystyle{IEEEtran}
\bibliography{bibliographyLocal}

\section*{APPENDIX}\label{Appendix}
\subsection{Transitioning from high dimensional problem (\ref{eq:8}) to a low dimensional problem (\ref{eq:9})} \label{Appendix:A}
Denote $p_i$ as the patch that fully contains the slice $s_i=\mathbf{D}_L\alpha_i$, and define a patch-layer $L_{\tilde{i}}$ as the set of non-overlapping patches taken from the image that contains the patch $p_i$. We can write the identity matrix as a sum of non-overlapping patch-extraction matrices $\sum_{k\in L_{\tilde{i}}}\mathbf{P}_k^T\mathbf{P}_k = \mathbf{I}$, where $\mathbf{P}_i$ is one of these matrices.  
By using these definitions and writing the definition of $R_i$ (the residual image without the contribution of the i-th needle) we can write the $l_2$ fidelity term of Equation (\ref{eq:8}) as  
\begin{equation}\nonumber
\begin{aligned}
\argmin_{\alpha_i}\frac{1}{2}\| \sum_{k\in L_{\tilde{i}}}\mathbf{P}_k^T\mathbf{P}_k R_i-\mathbf{P}_i^T\mathbf{D}_L\alpha_i\|_2^2 =~~~~~\\
\argmin_{\alpha_i}\frac{1}{2}\| \sum_{k\in L_{\tilde{i}}}\mathbf{P}_k^T\mathbf{P}_k (\signal-\sum_{\stackrel{j=1}{j\neq i}}^N\mathbf{P}_j^T\mathbf{D}_L\alpha_j)-\mathbf{P}_i^T\mathbf{D}_L\alpha_i\|_2^2=\\
\argmin_{\alpha_i}\frac{1}{2}\| \sum_{\stackrel{k\neq i}{k\in L_{\tilde{i}}}}\mathbf{P}_k^T\mathbf{P}_k (\signal-\sum_{\stackrel{j=1}{j\neq i}}^N\mathbf{P}_j^T\mathbf{D}_L\alpha_j)+ ~~~~~~~~~~~~\\
\mathbf{P}_i^T(\mathbf{P}_i(\signal-\sum_{\stackrel{j=1}{j\neq i}}^N\mathbf{P}_j^T\mathbf{D}_L\alpha_j)-\mathbf{D}_L\alpha_i)\|_2^2.
\end{aligned}
\end{equation}
Since the patch-extraction matrix $\mathbf{P}_i^T$ is orthogonal to all the matrices $\mathbf{P}_k^T$ for $ k\neq i$, the above is equal to 
\begin{equation}\nonumber
\begin{aligned}
\argmin_{\alpha_i}\frac{1}{2}\| \sum_{\stackrel{k\in L_{\tilde{i}}}{k\neq i}}\mathbf{P}_k^T\mathbf{P}_k (\signal-\sum_{\stackrel{j=1}{j\neq i}}^N\mathbf{P}_j^T\mathbf{D}_L\alpha_j)\|_2^2+ ~~~~~~~~~~~~\\
\|\mathbf{P}_i^T(\mathbf{P}_i(\signal-\sum_{\stackrel{j=1}{j\neq i}}^N\mathbf{P}_j^T\mathbf{D}_L\alpha_j)-\mathbf{D}_L\alpha_i)\|_2^2.\\
\end{aligned}
\end{equation}
Note that the first term of the above objective does not depend on $\alpha_i$, and thus we can ignore this term in our minimization of the objective:
\begin{equation}\nonumber
\begin{aligned}
\argmin_{\alpha_i}\frac{1}{2}\|\mathbf{P}_i^T(\mathbf{P}_i(\signal-\sum_{\stackrel{j=1}{j\neq i}}^N\mathbf{P}_j^T\mathbf{D}_L\alpha_j)-\mathbf{D}_L\alpha_i)\|_2^2.
\end{aligned}
\end{equation}
In addition, the matrix $\mathbf{P}_i^T$ translates a patch-size vector to the i-th position in the global vector padded with zeros. Hence, we can ignore all the zero-entires in the resulting vector, and the problem becomes equivalent to solving the following reduced minimization problem:
\begin{equation}\nonumber
\begin{aligned}
\argmin_{\alpha_i}\frac{1}{2}\|\mathbf{P}_i(\signal-\sum_{\stackrel{j=1}{j\neq i}}^N\mathbf{P}_j^T\mathbf{D}_L\alpha_j)-\mathbf{D}_L\alpha_i\|_2^2=\\
\argmin_{\alpha_i}\frac{1}{2}\|\mathbf{P}_iR_i-\mathbf{D}_L\alpha_i\|_2^2,
\end{aligned}
\end{equation}
where $R_i=\signal-\sum_{\stackrel{j=1}{j\neq i}}^N\mathbf{P}_j^T\mathbf{D}_L\alpha_j$.
%\section*{APPENDIX}%\label{Appendix:B}
\subsection{The gradient calculation w.r.t the local dictionary of the minimization problem (\ref{eq:10})}\label{Appendix:B}
We can rewrite $\mathbf{D}_L$ as a column vector $d_L$ and write Equation (\ref{eq:10}) as (we use $\otimes$ to denote the Kronecker matrix product \cite{minka2000old}):
\begin{equation}\nonumber
\argmin_{\mathbf{D}_L}\frac{1}{2}\|\signal-\sum_{i=1}^N\mathbf{P}_i^T(\alpha_i\otimes \mathbf{I}_{n\times n})d_L\|_2^2,
\end{equation}
and by defining $\mathbf{A}_i = (\alpha_i\otimes \mathbf{I}_{n\times n})$, the above problem can be rewritten as
\begin{equation}\nonumber
\argmin_{\mathbf{D}_L}\frac{1}{2}\|\signal-\sum_{i=1}^N\mathbf{P}_i^T\mathbf{A}_id_L\|_2^2.
\end{equation}
Now it easy to see that the gradient of the above problem w.r.t $d_L$ is given by
\begin{equation}\nonumber
\begin{aligned}
\nabla_{d_L} =-(\sum_{i=1}^N\mathbf{P}_i^T\mathbf{A}_i)^T(\signal-\sum_{i=1}^N\mathbf{P}_i^T\mathbf{A}_id_L)=\\
-\sum_{i=1}^N\mathbf{A}_i^T\mathbf{P}_i(\signal-\sum_{i=1}^N\mathbf{P}_i^T\mathbf{A}_id_L)=\\
-\sum_{i=1}^N\mathbf{A}_i^T\mathbf{P}_i(\signal-\widehat{\signal}).
\end{aligned}
\end{equation}
Note that $\mathbf{P}_i(\signal-\widehat{\signal})$ is the i-th patch of the residual image, and by substituting back the definition of $\mathbf{A}_i$, and using the same property as before (see \cite{minka2000old} property no. (40)) we can write the gradient as
\begin{equation}\nonumber
\begin{aligned}
\nabla_{d_L} =-\sum_{i=1}^N(\alpha_i\otimes \mathbf{I}_{n\times n})^T\mathbf{P}_i(\signal-\widehat{\signal})=\\
-\sum_{i=1}^N vec(\mathbf{P}_i(\signal-\widehat{\signal})\cdot \alpha_i^T),
\end{aligned}
\end{equation}
where $vec(\cdot)$ denotes the vec-operator that stacks the columns of the gradient matrix into a vector, so by reshaping the above expression we get the final expression for the gradient
\begin{equation}\nonumber
\nabla_{\mathbf{D}_L} =-\sum_{i=1}^N\mathbf{P}_i(\signal-\widehat{\signal})\cdotp\alpha_i^T.
\end{equation}

\begin{comment}
%\section*{APPENDIX C}%\label{Appendix:C}
\subsection{Close form solution of the MOD dictionary update rule} \label{Appendix:C}
By taking the above first expression for the gradient and equating it to zero we get a close form solution for the MOD update rule
\begin{equation}\nonumber
\begin{aligned}
d_L = [(\sum_{i=1}^N\mathbf{P}_i^T\mathbf{A}_i)^T(\sum_{i=1}^N\mathbf{P}_i^T\mathbf{A}_i)]^{-1}(\sum_{i=1}^N\mathbf{P}_i^T\mathbf{A}_i)^T \signal =\\
[\widetilde{\mathbf{A}}^T\widetilde{\mathbf{A}}]^{-1}\widetilde{\mathbf{A}}^T \signal,~~~~~~~~~~~~~~~~~~~~~
\end{aligned}
\end{equation}
where we have defined $\widetilde{\mathbf{A}} = \sum_{i=1}^N\mathbf{P}_i^T\mathbf{A}_i$. By reshaping the above expression we get the final update rule for the dictionary update:
\begin{equation}\nonumber
\mathbf{D}_L = \mathit{S}\{[\widetilde{\mathbf{A}}^T\widetilde{\mathbf{A}}]^{-1}\widetilde{\mathbf{A}}^T \signal\},
\end{equation}
where $\mathit{S}$ is the operator that reshape the vectorized form of the dictionary $d_L$ to its original size $n \times m$.
\end{comment}

\end{document}